\documentclass{article} 
\usepackage[preprint]{colm2026_conference}

\usepackage[T1]{fontenc}
\usepackage{amsmath}
\usepackage{microtype}
\usepackage{hyperref}
\usepackage{url}
\usepackage{booktabs}
\usepackage{multirow}
\usepackage{xcolor}
\usepackage{subcaption}
\usepackage{graphicx}
\usepackage{pdflscape}
\usepackage{caption}

\usepackage{lineno}

\definecolor{darkblue}{rgb}{0, 0, 0.5}
\hypersetup{colorlinks=true, citecolor=darkblue, linkcolor=darkblue, urlcolor=darkblue}

\graphicspath{{figures/}}

\newcommand{\Cref}[1]{Table~\ref{#1}}
\newcommand{\cref}[1]{Table~\ref{#1}}

\title{Do LLM Decoders Listen Fairly? \\ Benchmarking How Language Model Priors Shape Bias in Speech Recognition}

\makeatletter
\def\@maketitle{\vbox{\hsize\textwidth
{\centering\Large\bf \@title\par}
\vskip 0.25in
{\centering\@author\par}
\vskip 0.3in minus 0.1in}}
\makeatother

\author{%
Srishti Ginjala$^{1}$, Eric Fosler-Lussier$^{1}$, Christopher W. Myers$^{2}$, Srinivasan Parthasarathy$^{1}$ \\[0.4em]
$^{1}$The Ohio State University, Columbus, OH, USA \\
$^{2}$Air Force Research Laboratory, USA
}

\begin{document}

\ifcolmsubmission
\linenumbers
\fi

\maketitle
\lhead{}\chead{}\rhead{}
\renewcommand{\headrulewidth}{0pt}

\begin{abstract}
As pretrained large language models replace task-specific decoders in speech recognition, a critical question arises: do their text-derived priors make recognition fairer or more biased across demographic groups?
We evaluate nine models spanning three architectural generations (CTC with no language model, encoder-decoder with an implicit LM, and LLM-based with an explicit pretrained decoder) on ${\sim}$43{,}000 utterances across five demographic axes (ethnicity, accent, gender, age, first language) using Common Voice~24 and Meta's Fair-Speech, a controlled-prompt dataset that eliminates vocabulary confounds.
On clean audio, three findings challenge assumptions: LLM decoders do not amplify racial bias, as the most ethnicity-fair competitive model (Granite-8B, max/min WER ratio\,{=}\,2.28) uses an explicit LLM decoder; Whisper's implicit-LM decoder exhibits pathological hallucination on Indian-accented speech with a non-monotonic insertion-rate spike to 9.62\% at large-v3; and audio compression predicts accent fairness more than LLM scale.
We then stress-test these findings under 12 acoustic degradation conditions (noise, reverberation, silence injection, chunk masking) across both datasets, totaling 216 inference runs.
Severe degradation paradoxically \emph{compresses} fairness gaps as all groups converge to high WER, but silence injection \emph{amplifies} Whisper's accent bias up to 4.64$\times$ by triggering demographic-selective hallucination.
Under masking, Whisper enters catastrophic repetition loops (86\% of 51{,}797 insertions) while explicit-LLM decoders produce 38$\times$ fewer insertions with near-zero repetition; high-compression audio encoding (Q-former) reintroduces repetition pathology even in LLM decoders.
These results suggest that audio encoder design, not LLM scaling, is the primary lever for equitable and robust speech recognition.
\end{abstract}

\section{Introduction}
\label{sec:intro}

Language models are increasingly forming the decoder backbone of speech recognition systems.
Three architectural generations now coexist: CTC-based systems that use no language model, mapping audio directly to characters; encoder-decoder systems like Whisper that learn an \emph{implicit} language model from their training transcriptions; and the newest generation, which routes audio embeddings into a \emph{pretrained} LLM whose distributional priors are inferred from text-only corpora.
Each generation integrates language priors more deeply into recognition.
But does a stronger language prior help all speakers equally, or does it encode the biases of its predominantly standard-English training text?

ASR is known to exhibit demographic biases: \citet{koenecke2020racial} showed that five commercial systems produced roughly double the error rate for Black speakers compared to White speakers, and \citet{tatman2017gender} documented gender and dialect disparities.
These findings predate the LLM-decoder era.
A natural concern follows: pretrained LLMs, whose priors reflect standard written English, may amplify existing gaps when decoding non-standard or accented speech.
Moreover, real-world speech is rarely clean: noise, reverberation, and signal dropout force the decoder to compensate, moving it away from its language prior, potentially amplifying biases invisible on clean benchmarks.
This concern is urgent because LLM-based ASR is deployed at scale. Yet, no study has systematically tested whether adding a pretrained language model to the decoder makes recognition more or less equitable, either on clean speech or under acoustic degradation.

To our knowledge, we present the first study isolating how the degree of language model integration affects ASR fairness on clean speech and under controlled degradation.
We evaluate nine models (one CTC system with no LM, three encoder-decoder systems with an implicit LM, and five LLM-decoder systems with an explicit LM) on ${\sim}$43{,}000 utterances from Common Voice~24 \citep{ardila2020commonvoice} and Meta's Fair-Speech \citep{veliche2024fairspeech} across five demographic axes.
We test all models under 12 perturbation conditions (noise, reverb, silence, masking) on both datasets, totaling 216 inference runs.
Our contributions:
(1)~the first systematic study of LLM decoders' effect on ASR fairness across five demographic axes and 12 degradation conditions;
(2)~evidence that LLM decoders do \emph{not} amplify racial bias (\S\ref{sec:ethnicity});
(3)~identification of Whisper's pathological hallucination on Indian-accented speech (9.62\% insertion rate) while all LLM decoders remain below 3.1\% (\S\ref{sec:accent});
(4)~evidence that audio compression predicts accent fairness more than LLM scale (\S\ref{sec:accent});
(5)~the finding that severe degradation paradoxically compresses fairness gaps, with one critical exception: silence amplifies Whisper's accent bias up to 4.64$\times$ (\S\ref{sec:perturbation});
and (6)~demonstration that hallucination type under degradation is architecture-dependent, with high audio compression reintroducing pathological repetition even in LLM decoders (\S\ref{sec:perturbation:hallucination}).

\section{Benchmarking Setup}
\label{sec:models}

\subsection{Models: three generations of language model integration}
\label{sec:models:arch}

\Cref{tab:models} summarizes the nine models in our study.
\textbf{Generation~1}: Wav2Vec2-large \citep{baevski2020wav2vec}, a CTC encoder fine-tuned on LibriSpeech 960h, maps audio frames to character probabilities with no autoregressive decoder and no language model.
\textbf{Generation~2}: three Whisper checkpoints \citep{radford2023whisper} (small 244M, medium 764M, large-v3 1.5B), whose autoregressive decoder learns an implicit language model from 680{,}000 hours of paired transcriptions.
\textbf{Generation~3}: five models that route audio embeddings into a pretrained LLM backbone: Qwen3-ASR (0.6B, 1.7B) with low-compression direct audio-token projection \citep{shi2026qwen3}; Canary-Qwen-2.5B with medium compression via FastConformer \citep{nvidia2025canary}; and Granite-Speech (2B, 8B) with high compression through a Q-former bottleneck sharing one encoder across both sizes \citep{saon2025granite}.
This yields controlled comparisons: Whisper small$\to$medium$\to$large-v3 (implicit-LM scaling); Qwen3 0.6B$\to$1.7B (explicit-LLM, low compression); Granite 2B$\to$8B (explicit-LLM, constant high compression); and Qwen3-1.7B vs.\ Granite-2B ($\sim$same parameters, different compression). We selected these nine open-weight models from HuggingFace as representative of each architectural generation based on Open ASR Leaderboard performance (as of February 1,2026) and community adoption, subject to our computational budget.

\begin{table}[t]
\centering
\caption{The nine ASR models evaluated, grouped by the role of the language model in decoding. \emph{Audio compression} refers to how the raw waveform is transformed before reaching the decoder. Overall WER (\%) on each evaluation corpus is shown at right.}
\label{tab:models}
\resizebox{\textwidth}{!}{
\begin{tabular}{llclcllrrr}
\toprule
\# & Model & Gen & Architecture & Params & Audio Compression & LM Type & LS & CV & FS \\
\midrule
1 & Wav2Vec2-large & 1 & CTC encoder & 317M & None (raw frames) & No LM & 1.79 & 22.72 & 32.15 \\
\midrule
2 & Whisper-small & 2 & Enc-Dec Transformer & 244M & Log-mel 80d & Implicit LM & 3.50 & 16.59 & 11.51 \\
3 & Whisper-medium & 2 & Enc-Dec Transformer & 764M & Log-mel 80d & Implicit LM & 2.99 & 12.59 & 8.75 \\
4 & Whisper-large-v3 & 2 & Enc-Dec Transformer & 1.5B & Log-mel 128d & Implicit LM & 1.92 & 10.96 & 7.79 \\
\midrule
5 & Qwen3-ASR-0.6B & 3 & Audio enc + Qwen3 & 0.6B & Low (direct) & Explicit LLM & 2.13 & 10.08 & 5.89 \\
6 & Qwen3-ASR-1.7B & 3 & Audio enc + Qwen3 & 1.7B & Low (direct) & Explicit LLM & 1.60 & 7.76 & 4.73 \\
7 & Canary-Qwen-2.5B & 3 & FastConformer + Qwen-2.5B & 2.5B & Medium & Explicit LLM & 1.61 & 7.72 & 6.60 \\
8 & Granite-Speech-2B & 3 & Conformer + Q-former + LLM & 2.0B & High (Q-former) & Explicit LLM & 1.53 & 10.09 & 8.99 \\
9 & Granite-Speech-8B & 3 & Conformer + Q-former + LLM & 8.0B & High (Q-former) & Explicit LLM & 2.42 & 10.86 & 8.04 \\
\bottomrule
\end{tabular}
}
\end{table}

\subsection{Datasets}
\label{sec:models:data}

We evaluate on three English speech corpora.
\textbf{Common Voice~24 test split} \citep{ardila2020commonvoice} provides ${\sim}$16{,}400 crowd-sourced read-speech utterances with self-reported accent (6 groups, $n \geq 50$), gender, and age (5 bins); its large Indian-accent subgroup ($n{=}511$) enables fine-grained hallucination analysis.
\textbf{Fair-Speech} \citep{veliche2024fairspeech} contains ${\sim}$26{,}470 utterances from U.S.\ English speakers who all read the \emph{same prompted sentences}, eliminating vocabulary confounds; any WER differences are attributable to acoustic or model factors.
It provides ethnicity (7 groups), gender, age, first language (30+ L1s), and socioeconomic status (3 levels).
To our knowledge, we report the first comprehensive multi-model benchmark on Fair-Speech.
\textbf{LibriSpeech test-clean} \citep{panayotov2015librispeech} (2{,}620 utterances, no demographics) serves as a reference baseline.

\subsection{Metrics}
\label{sec:models:metrics}

We report word error rate (WER) as the primary measure of accuracy and \textbf{max/min ratio} (MMR) as the fairness metric: $MMR = \max_{g \in G} \text{WER}_g \,/\, \min_{g \in G} \text{WER}_g$ over groups with $n \geq 50$.
MMR\,{=}\,1.0 indicates perfect parity; we prefer it over absolute gap because it is scale-invariant.
All group-level WERs include 95\% bootstrap CIs (200 resamples).
Following \citet{morris2004wer}, we decompose errors into substitutions, insertions, and deletions to distinguish acoustic confusion from pathological decoder-driven errors, such as hallucinations.

For perturbation experiments, we additionally report the \textbf{fairness gap amplification ratio}: $\alpha = MMR^{pert} / MMR^{clean}$, where $\alpha > 1$ indicates degradation widens demographic gaps and $\alpha < 1$ indicates compression.
We classify insertions into \textbf{repetition} (autoregressive loops), \textbf{syntactic} (function words), and \textbf{content} (semantic fabrications) to characterize architecture-specific hallucination behavior under degradation.

\subsection{Perturbation Design Choices}
\label{sec:models:perturbation}

Real-world audio is rarely clean; degraded signals force the decoder to compensate from its language prior, potentially revealing demographic biases invisible on clean benchmarks.
We test all nine models under four perturbation types at three severity levels:
\textbf{additive noise} (SNR 20/10/0\,dB; source: MUSAN noise corpus \citep{snyder2015musan});
\textbf{reverberation} (RT60 0.3/0.6/1.0\,s; source: OpenSLR RIRs \citep{ko2017study});
\textbf{silence injection} (25/50/75\% of utterance duration inserted at a random position); and
\textbf{chunk masking} (10/20/30\% of audio zeroed out across multiple chunks).
Noise and reverb degrade signal quality while preserving temporal structure; silence and masking remove acoustic information entirely, maximally forcing decoder reliance on its language prior.
We deliberately exclude standard non-destructive augmentations (e.g., pitch shifting, speed scaling, filtering) because comprehensive robustness to these has already been studied by \citet{shah2025robustbench}. Moreover, unlike signal dropout, these continuous transformations do not differentially stress the decoder's language model prior, which is the core focus of our analysis.
Perturbed audio is generated offline; inference configuration matches the clean evaluation exactly (Appendix~\ref{app:inference}).
In total 9 models $\times$ 12 conditions $\times$ 2 datasets = 216 inference runs.

\section{Clean evaluation}
\label{sec:clean}

\subsection{Ethnicity fairness}
\label{sec:ethnicity}
\label{sec:ethnicity:overall}
\label{sec:ethnicity:wer}

\noindent {\bf Baseline performance:} All 9 models achieve 1.5-3.5\% WER on LibriSpeech test-clean (Table~\ref{tab:models}), confirming base competence. Performance diverges on diverse speech: on Fair-Speech, Gen~3 models lead (Qwen3-1.7B: 4.73\%), Gen~2 models range from 7.8-11.5\%, and Wav2Vec2-large reaches 32.15\%, an 18-fold increase over its LibriSpeech WER reflecting a domain mismatch between its training data and Fair-Speech's mobile-recorded speech.

\noindent {\bf WER by Ethnicity:} Figure~\ref{fig:heatmaps}a presents WER by ethnicity on Fair-Speech (full table in Appendix~\ref{app:ethnicity}).
The most consistent finding in this study is that \textbf{Black/African-American speakers have the highest WER across all nine models}, regardless of generation or architecture.
This holds for Gen~1 (Wav2Vec2: 41.6\%), Gen~2 (Whisper-large-v3: 10.2\%), and Gen~3 (Qwen3-1.7B: 8.5\%).
The gap is statistically significant for 8 of 9 models: bootstrap 95\% CIs for Black/AA and White WER do not overlap (Table~\ref{tab:ci_ethnicity}).
The sole exception is Whisper-large-v3, where Black/AA 10.25\% [9.61, 10.81] and White 9.42\% [8.21, 10.64] overlap, where Whisper's scaling genuinely narrows the ethnicity gap, even as it simultaneously worsens accent fairness through hallucination (\S\ref{sec:accent}).
This underscores a challenge with {\it intersectional fairness} where improvements on one sensitive axis often degrades another.

\label{sec:ethnicity:verdict}
\noindent {\bf Pre-trained LLM priors do not widen racial gaps.} Contrary to the above hypothesis, Gen~3 models do \emph{not} exhibit worse ethnicity fairness than Gen~2.
Among competitive models, Granite-8B (Gen~3, MMR\,{=}\,2.28) achieves the best ethnicity fairness, outperforming all three Whisper models (MMR 3.13-4.04; Table~\ref{tab:fs_wer_ethnicity}).
Gender disparities are negligible on Common Voice (max MMR\,{=}\,1.12) but substantial on Fair-Speech (Qwen3-0.6B gender MMR\,{=}\,2.42), likely due to Fair-Speech's controlled prompts revealing acoustic-level effects masked by Common Voice's heterogeneous sentences (Appendix~\ref{app:gender}).

\label{sec:ethnicity:paradox}
\noindent {\bf A measurement paradox emerges:} Qwen3-1.7B achieves the \emph{lowest absolute} Black/AA WER (8.45\%) yet the \emph{highest relative gap} (+203\%) because its White WER is extremely low (2.79\%).
Conversely, Wav2Vec2 achieves the smallest relative gap (MMR\,{=}\,1.58), but this is merely ``low-accuracy parity'' where all groups suffer unacceptably high error rates (e.g., 41.6\% Black/AA).
A recommendation here is that practitioners should report both absolute WER per group and relative disparity metrics.

\begin{figure}[t]
\centering
\begin{subfigure}[t]{0.49\textwidth}
  \centering
  \includegraphics[width=\textwidth]{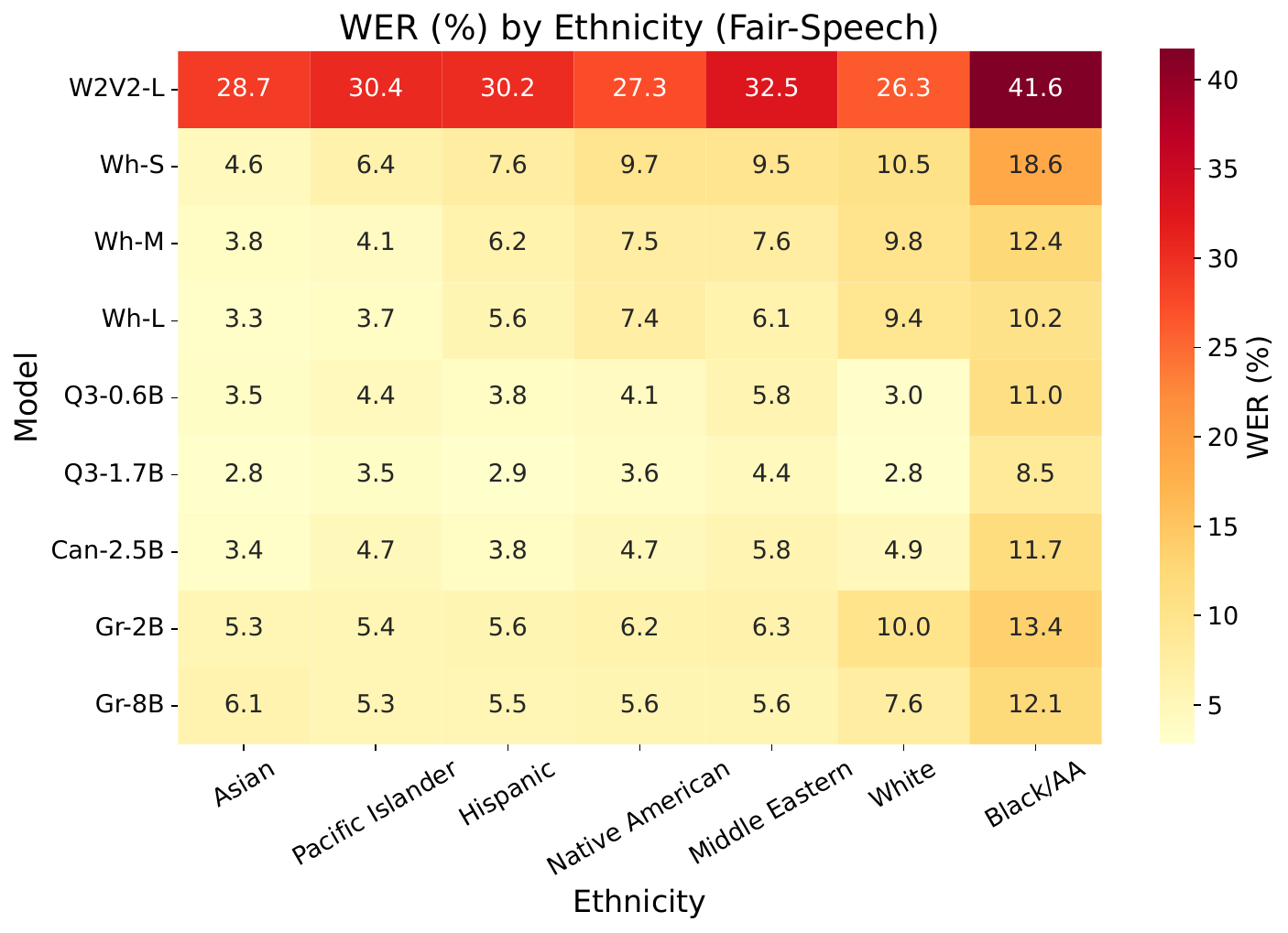}
  \caption{Ethnicity WER on Fair-Speech.}
  \label{fig:heatmap_ethnicity}
\end{subfigure}
\hfill
\begin{subfigure}[t]{0.49\textwidth}
  \centering
  \includegraphics[width=\textwidth]{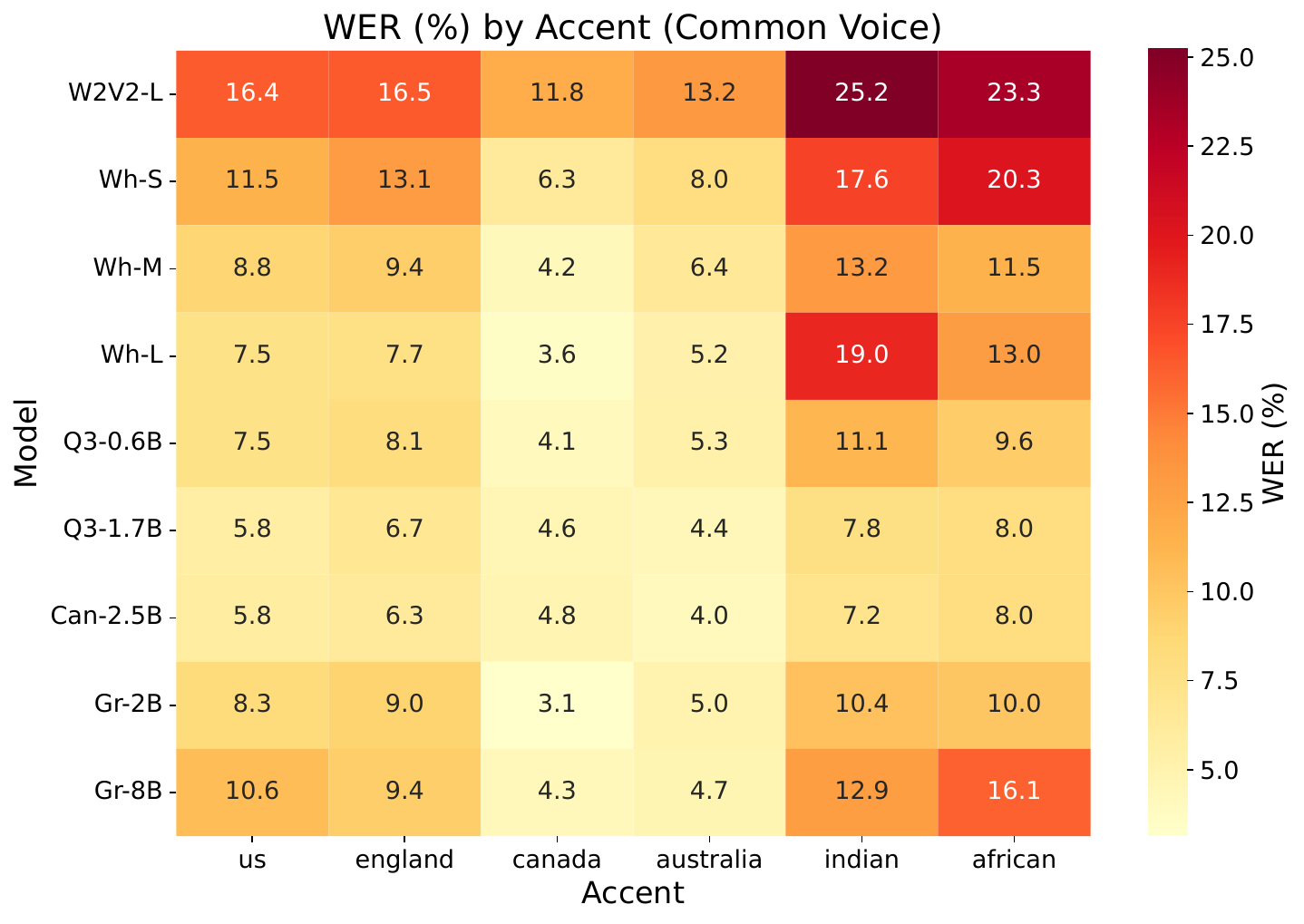}
  \caption{Accent WER on Common Voice~24.}
  \label{fig:heatmap_accent}
\end{subfigure}
\caption{WER (\%) by (a) ethnicity and (b) accent. Black/AA speakers face the highest WER everywhere. Indian/African accents are hardest, with Whisper-large-v3 underperforming small on Indian accent due to hallucination.}
\label{fig:heatmaps}
\end{figure}

\subsection{Accent fairness and decoder hallucination}
\label{sec:accent}
\label{sec:accent:wer}

\noindent {\bf Baseline performance:} Figure~\ref{fig:heatmaps}b presents WER across six accent groups on Common Voice~24 (full table in Appendix~\ref{app:accent}).
Indian and African accents are hardest for every model, but error sources differ dramatically by architecture.
Qwen3-1.7B is the most equitable (accent MMR\,{=}\,1.82, range 4.41-8.03\%), while Whisper-large-v3 is the least equitable (MMR\,{=}\,5.34) despite its best overall WER.
{\it The surprising paradox:} Whisper-large-v3's Indian-accent WER (19.0\%) is \emph{worse} than Whisper-small's (17.6\%).

\label{sec:accent:compression}
\noindent {\bf Compression predicts accent fairness.}
Among Gen~3 models at comparable scale ($\approx$2B), low-compression Qwen3-1.7B (accent MMR\,{=}\,1.82) outperforms medium-compression Canary-2.5B (2.00) and high-compression Granite-2B (3.30).
This dynamically shifts for ethnicity: high-compression Granite-2B (2.53) outperforms low-compression Qwen3-1.7B (3.03), suggesting accent depends on fine-grained phonetic and prosodic cues that are easily lost in compression while ethnicity-relevant patterns may be recovered despite compression.

\label{sec:accent:hallucination}
\noindent {\bf Whisper hallucinates on Indian-accented speech.}
Table~\ref{tab:whisper_scaling} shows that Whisper's insertion rate on Indian accent ($n{=}511$) follows a non-monotonic trajectory: small 3.22\% $\to$ medium 1.53\% $\to$ large-v3 \textbf{9.62\%}.
At large-v3, insertions become the dominant error type (50.7\% of all errors).
All five Gen~3 models remain below 3.1\% on the same utterances (Appendix~\ref{app:insertion}).
This pathology is accent-selective: Whisper-large-v3's insertion rates on US (0.96\%) and Canadian (0.52\%) accents are comparable to other models.

\begin{table}[t]
\centering
\caption{Whisper scaling on Indian-accented speech (Common Voice 24, $n{=}511$).
  Insertion rate is non-monotonic: medium \emph{improves} over small, then large-v3 spikes pathologically.}
\label{tab:whisper_scaling}
\begin{tabular}{lccccr}
\toprule
Model & Params & Indian WER & Ins.\ Rate & Sub.\ Rate & Ins.\ \% of Errors \\
\midrule
Whisper-small   & 244M & 17.6\% & 3.22\% & 12.92\% & 18.3\% \\
Whisper-medium  & 764M & 13.2\% & 1.53\% & 9.99\%  & 11.6\% \\
Whisper-large-v3 & 1.5B & 19.0\% & \textbf{9.62\%} & 8.32\%  & \textbf{50.7\%} \\
\bottomrule
\end{tabular}
\end{table}

\begin{figure}[t]
\centering
\includegraphics[width=0.8\textwidth]{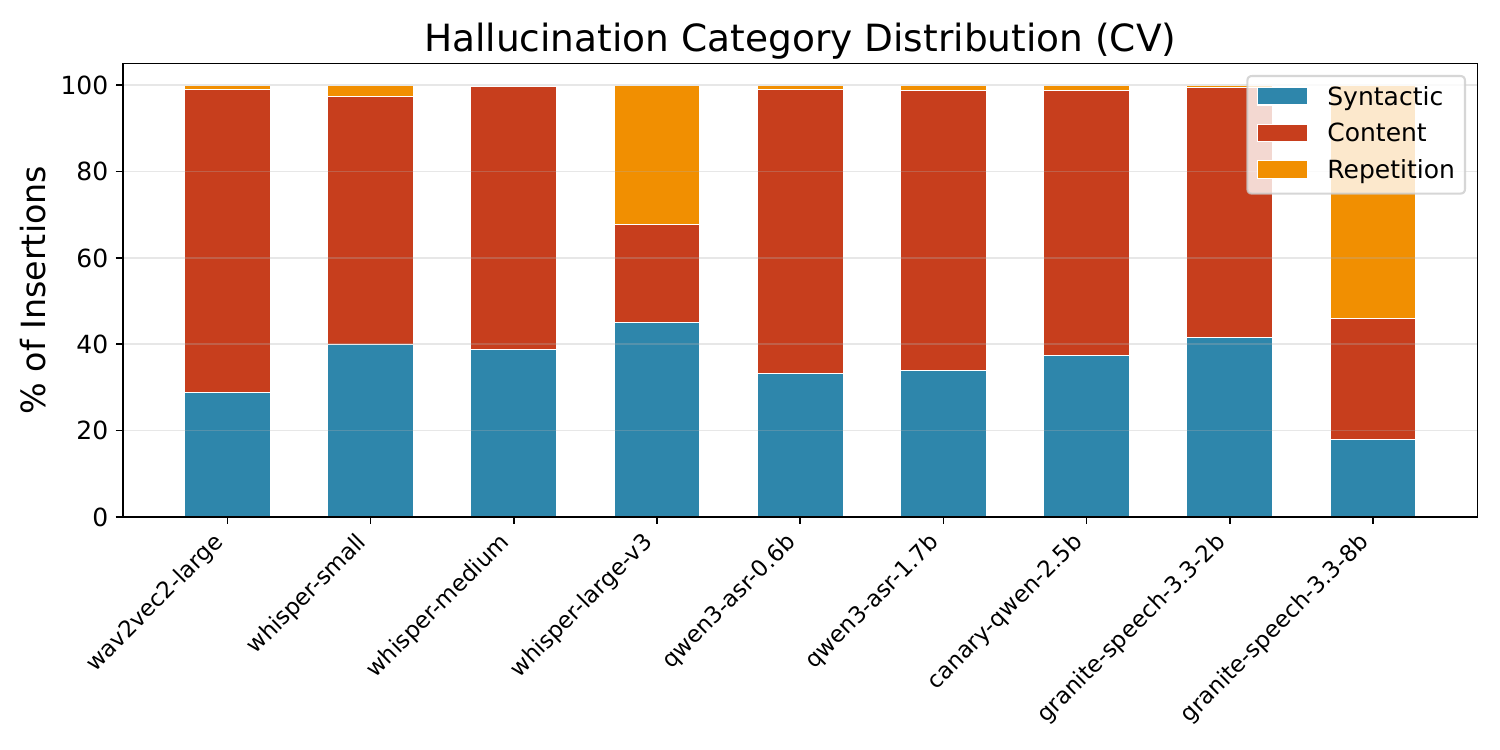}
\caption{Hallucination categories on Common Voice~24. Whisper-large-v3 is driven by repetition loops. Granite's repetition pathology is diagnosed under signal dropout.}
\label{fig:hallucination_categories_cv}
\end{figure}

\label{sec:accent:type}
At the 9.62\% insertion rate from Table~\ref{tab:whisper_scaling}, Whisper-large-v3 produces 495 insertions on the 511 Indian-accent utterances. Table~\ref{tab:indian_hallucination} (Appendix~\ref{app:insertion}) decomposes these by type: 43\% are repetition loops, 48\% syntactic completions, and 9.3\%
  content hallucinations, semantic fabrications that alter meaning (e.g., ``grow
  sea monkeys'' $\to$ ``go and see what happens'';
  Table~\ref{tab:hallucination_examples}).
  Figure~\ref{fig:hallucination_categories_cv} shows the all-accent category
  distribution. Gen~3 models' far rarer insertions are predominantly benign function words
  (Table~\ref{tab:indian_hallucination}).

\begin{table}[b]
\centering
\caption{Hallucinations on Indian-accented speech. Whisper-large-v3 fabricates semantic content; Canary-Qwen-2.5B inserts function words. Insertions \textbf{bolded}.}
\label{tab:hallucination_examples}
\resizebox{\textwidth}{!}{
\begin{tabular}{lp{5.5cm}p{5.5cm}l}
\toprule
Model & Reference & Hypothesis & Category \\
\midrule
Whisper-large-v3
  & perhaps you should just grow sea monkeys
  & perhaps \textbf{we} should just \textbf{go and see what happens}
  & Content \\
Whisper-large-v3
  & playing the 3 year old Kimberley Macdonald
  & playing the 3 year old \textbf{kim ming mcdonnell}
  & Content \\
Whisper-large-v3
  & boyd was not always popular with rank and file officers
  & \textbf{the 1st day of the 1st day of the 1st day of\ldots}
  & Repetition \\
\midrule
Canary-Qwen-2.5B
  & in korea carnations express admiration love and gratitude
  & in korea \textbf{the} carnations express admiration love and gratitude
  & Syntactic \\
Canary-Qwen-2.5B
  & he is noted for his studies of plants in the south of africa
  & he is noted for his studies of plants in \textbf{in} the south of africa
  & Syntactic \\
\bottomrule
\end{tabular}
}
\end{table}

\subsection{Scaling trajectories}
\label{sec:scaling}

\begin{figure}[ht]
\centering
\includegraphics[width=0.8\textwidth]{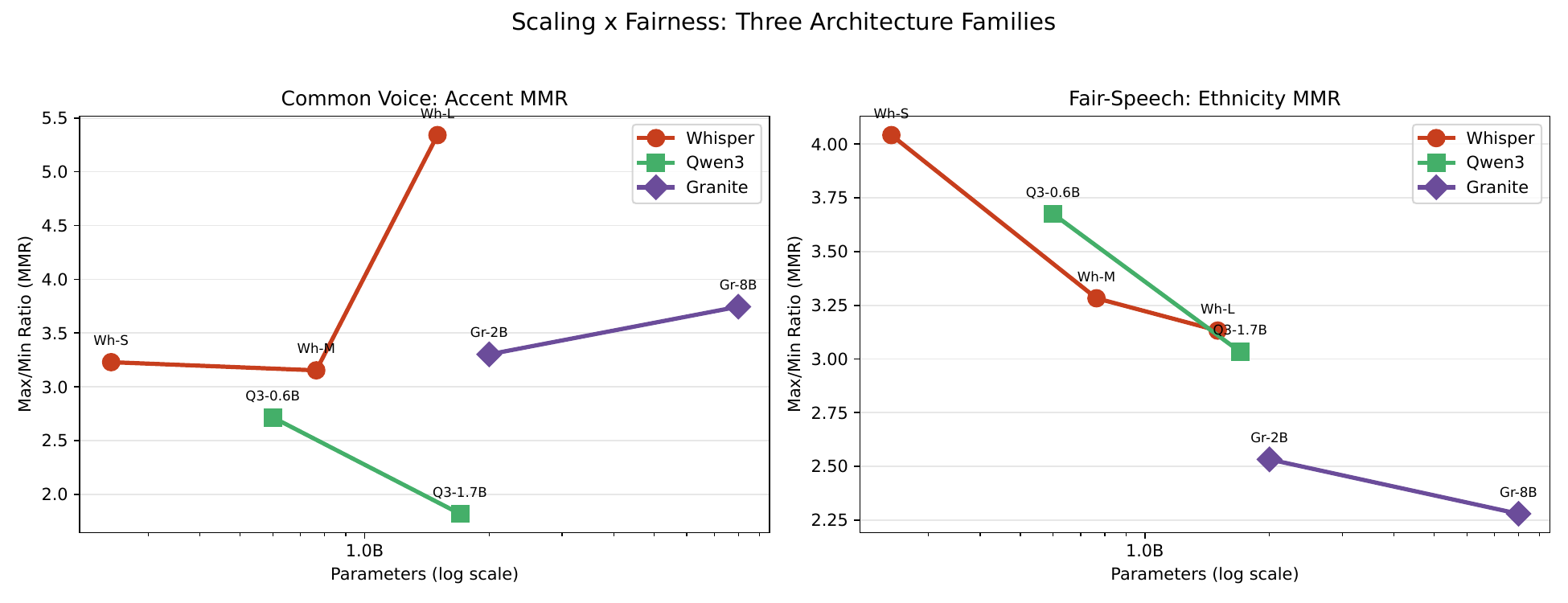}
\caption{Scaling trajectories. Qwen3 improves both accuracy and fairness. Whisper worsens accent fairness via hallucination. Granite shows dataset-dependent effects.}
\label{fig:scaling_curves}
\end{figure}

\noindent {\bf Scaling effects.} Scaling effects differ by architecture (Figure~\ref{fig:scaling_curves}).
Qwen3 (0.6B$\to$1.7B, low compression) improves both accuracy and fairness: accent MMR drops from 2.72 to 1.82 and ethnicity MMR from 3.68 to 3.03.
Whisper scaling improves ethnicity fairness (MMR: 4.04$\to$3.28$\to$3.13) but \emph{worsens} accent fairness (MMR: 3.23$\to$3.15$\to$5.34) due to the Indian-accent hallucination pathology (\S\ref{sec:accent:hallucination}).
Granite (2B$\to$8B, high compression) shows dataset-dependent effects: ethnicity fairness improves on Fair-Speech (MMR: 2.53$\to$2.28) but accent fairness worsens on Common Voice (MMR: 3.30$\to$3.74), suggesting the Q-former bottleneck limits the benefit of additional LLM capacity for acoustically diverse speech.

Additional demographic analyses (first language, age, and socioeconomic status) appear in Appendices~\ref{app:l1}-\ref{app:ses}.
{\it Key findings:} Whisper's implicit LM creates extreme L1 sensitivity (L1 MMR up to 15.05; Gen~3 $\leq$ 5.38); age$\times$architecture interactions flip which group is worst; SES gaps are moderate (MMR 1.13-2.17).

\section{Fairness under acoustic degradation}
\label{sec:perturbation}

Having established clean-audio fairness patterns, we now test whether they hold when acoustic degradation forces the decoder to increasingly rely on its language prior.

\subsection{Architecture-specific robustness}
\label{sec:perturbation:robustness}

\begin{figure}[ht]
\centering
\includegraphics[width=0.8\textwidth]{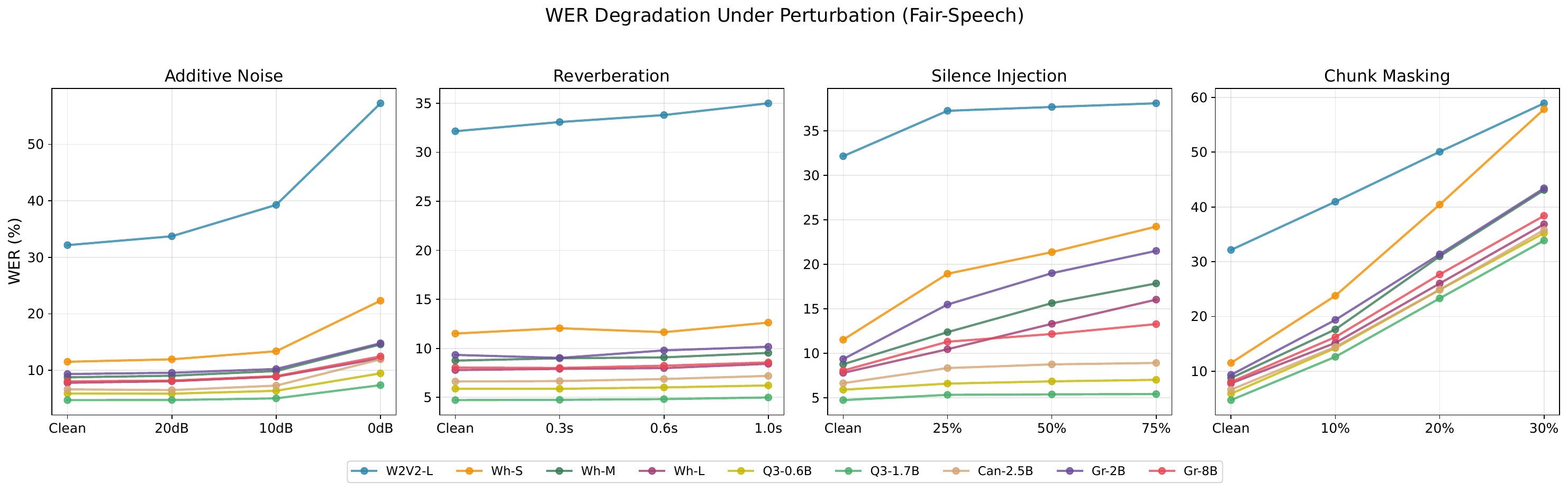}
\caption{WER degradation curves on Fair-Speech. Masking produces the steepest degradation. Qwen3-1.7B (green) is the most robust model.}
\label{fig:wer_curves_fs}
\end{figure}

\noindent {\bf Degradation robustness across ASR generations.} The three ASR generations differ markedly in degradation robustness (Figure~\ref{fig:wer_curves_fs}).
Under 30\% chunk masking on Fair-Speech, Wav2Vec2-large (no LM) reaches 58.9\% WER, nearly double its clean 32.1\%. Among Gen~3 models sharing Qwen LLM backbones, compression level predicts the degradation penalty: low-compression Qwen3-1.7B degrades to 33.9\%, while medium-compression Canary degrades worse to 35.8\%-a consistent trend observed across all four perturbation types.
Whisper spans a wide range: small (57.8\%) degrades comparably to Wav2Vec2, while large-v3 (36.8\%) approaches Gen~3 performance.
Highly-compressed Granite-2B (43.4\%) and Granite-8B (38.4\%) degrade much closer to Whisper. This confirms that audio bottleneck compression-not solely the LLM text pretraining data-dictates acoustic robustness.

\noindent {\bf Impact of perturbation type on robustness.} Reverberation is the mildest perturbation for models with language priors: at RT60\,{=}\,1.0\,s, Qwen3-1.7B barely increases (4.7\%$\to$5.0\%) and Whisper-large-v3 rises to only 8.4\%, while Wav2Vec2 reaches 35.0\%.
Additive noise at SNR\,{=}\,0\,dB falls between, with Qwen3-1.7B (7.4\%) and Whisper-large-v3 (12.1\%) demonstrating that stronger language priors better compensate for noise. Under silence injection, WER degrades minimally for Qwen3 and Canary, but increases almost linearly with silence duration for Whisper and Granite.
These patterns hold on Common Voice (Figure~\ref{fig:wer_curves_cv} in Appendix~\ref{app:perturbation_wer}): under 30\% masking, Qwen3-1.7B (35.7\%) outperforms all Whisper checkpoints (42.4-63.0\%) and Wav2Vec2 (57.6\%), while silence injection uniquely triggers demographic-selective hallucination (\S\ref{sec:perturbation:amplification}).

\subsection{Fairness gap amplification}
\label{sec:perturbation:amplification}

\begin{figure}[t]
\centering
\begin{subfigure}[t]{\textwidth}
  \centering
  \includegraphics[width=\textwidth]{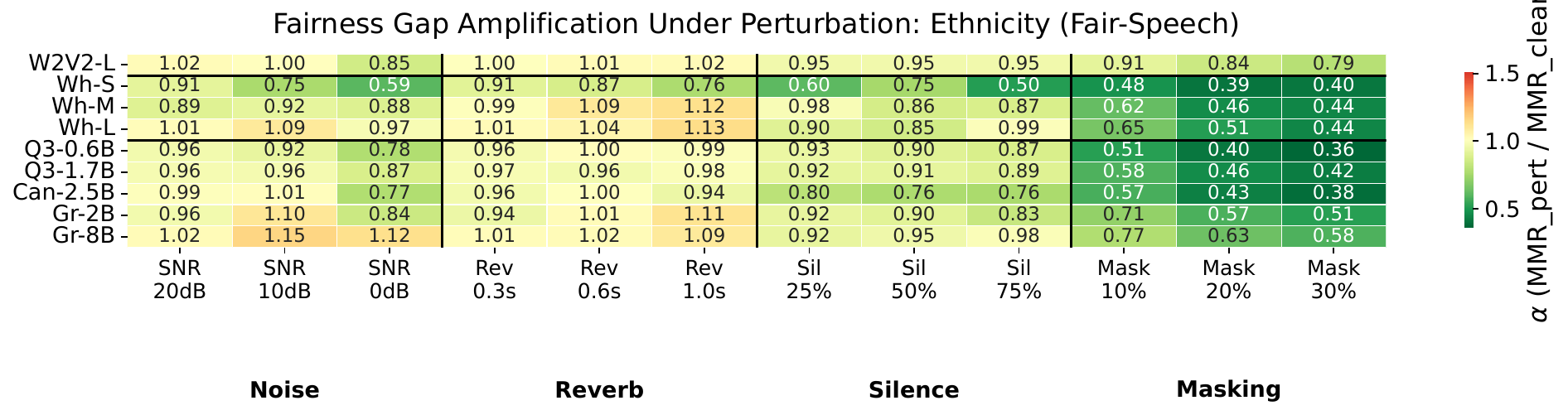}
  \caption{Ethnicity Gap Amplification (Fair-Speech)}
  \label{fig:ampl_heatmap_ethnicity}
\end{subfigure}
\par\bigskip
\begin{subfigure}[t]{\textwidth}
  \centering
  \includegraphics[width=\textwidth]{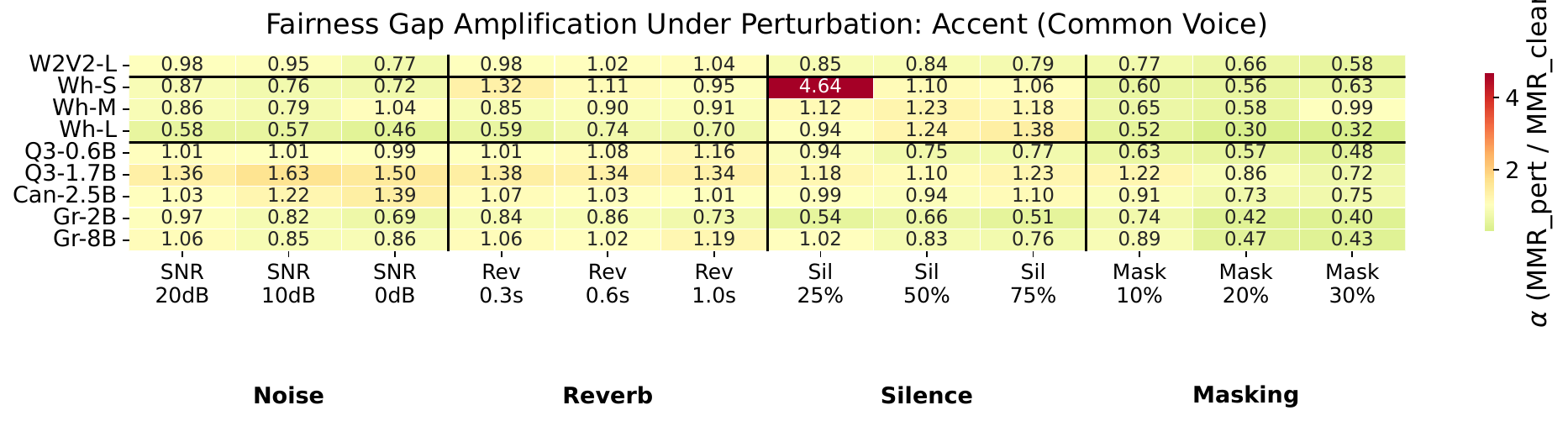}
  \caption{Accent Gap Amplification (Common Voice~24)}
  \label{fig:ampl_heatmap_accent}
\end{subfigure}
\caption{Fairness gap amplification ($\alpha$) by (a) ethnicity and (b) accent. Blue ($\alpha < 1$): compresses gaps; red ($\alpha > 1$): widens gaps. Masking universally compresses gaps. Noise amplifies Qwen3-1.7B bias ($\alpha$\,{=}\,1.63); silence redistributes Whisper-large-v3 bias (\S\ref{sec:perturbation:amplification}).}
\label{fig:amplification}
\end{figure}

Figure~\ref{fig:amplification} presents the fairness gap amplification ratio ($\alpha = MMR^{pert}/MMR^{clean}$) for ethnicity on Fair-Speech (panel a) and accent on Common Voice (panel b) across all nine models and 12 perturbation conditions.

\paragraph{Severe masking compresses fairness gaps.}
Under 30\% masking on Fair-Speech, \emph{every} model's ethnicity MMR compresses ($\alpha$\,{=}\,0.36-0.79; clean MMR 1.58-4.04, Table~\ref{tab:fs_wer_ethnicity}).
This mirrors the ``low-accuracy parity'' from \S\ref{sec:ethnicity}: when all groups suffer similarly catastrophic error rates, relative gaps shrink.
Accent fairness on Common Voice compresses similarly under masking ($\alpha$\,{=}\,0.32-0.99; Figure~\ref{fig:amplification}b).
At harshest noise (SNR\,{=}\,0\,dB), most models also compress ethnicity gaps ($\alpha$\,{=}\,0.59-0.97), though Granite-8B \emph{amplifies} ($\alpha$\,{=}\,1.12).
Reverb at RT60\,{=}\,1.0\,s produces mild ethnicity amplification for Whisper-large-v3 ($\alpha$\,{=}\,1.13), Whisper-medium (1.12), Granite-2B (1.11), and Granite-8B (1.09), remaining close to unity.

\paragraph{Accent gaps amplify under noise and silence.}
On Common Voice accent fairness (Figure~\ref{fig:amplification}b), amplification patterns are more prominent.
Qwen3-1.7B, the most equitable model on clean audio (accent MMR\,{=}\,1.82), shows the strongest amplification: $\alpha$\,{=}\,1.63 at SNR\,{=}\,10\,dB (MMR: 1.82\,$\to$\,2.96), with Indian-accent WER rising from 7.8\% to 11.3\% while Canadian barely changes (4.6\%\,$\to$\,3.9\%; Table~\ref{tab:accent_perturbation}).
Canary-2.5B amplifies similarly under noise ($\alpha$\,{=}\,1.39 at SNR\,{=}\,0\,dB).
Equitable clean performance does not guarantee equitable degradation.
Whisper-small exhibits the most extreme amplification: $\alpha$\,{=}\,4.64 at 25\% silence (MMR: 3.23\,$\to$\,15.0), driven by African-accent WER increasing 4.2$\times$ while Canadian barely changes. This effect is non-monotonic, collapsing to $\alpha$\,$\approx$\,1.1 at 50-75\% silence, suggesting a threshold where mild silence triggers catastrophic hallucination loops for specific accent groups.
Whisper-large-v3 under 75\% silence also amplifies accent bias ($\alpha$\,{=}\,1.38; MMR: 5.38\,$\to$\,7.45), but the mechanism is unexpected: Indian-accent WER \emph{decreases} from 19.2\% to 14.5\% as silence disrupts the hallucination pathway from \S\ref{sec:accent:hallucination}, while England-accent WER quadruples from 7.7\% to 30.4\%.
Perturbation \emph{redistributes} decoder bias across demographic groups rather than uniformly amplifying it.

\subsection{Hallucination under signal dropout}
\label{sec:perturbation:hallucination}

Both masking and silence injection, limits the reliance on acoustic information, forcing the decoder to generate text from its prior alone.
The resulting hallucination patterns reveal fundamental architectural differences.
Under 30\% masking on Fair-Speech, Whisper-small produces \textbf{51{,}797 insertions}, of which 86\% are repetition loops: catastrophic autoregressive cycling where the decoder repeats phrases hundreds of times.
Whisper-large-v3 is more contained (10{,}365 insertions, 35\% repetition) but still repetition-prone.
By contrast, Qwen3-1.7B produces only \textbf{1{,}353 insertions} (38$\times$ fewer than Whisper-small) with just 3\% repetition; its rare insertions are primarily syntactic completions (54\%) and content substitutions (44\%).
Wav2Vec2 (no LM) produces content-classified insertions (56\%, phonetic garble) with $<$2\% repetition, strongly suggesting that repetition loops are a decoder-driven phenomenon.

\begin{figure}[ht]
\centering
\includegraphics[width=0.8\textwidth]{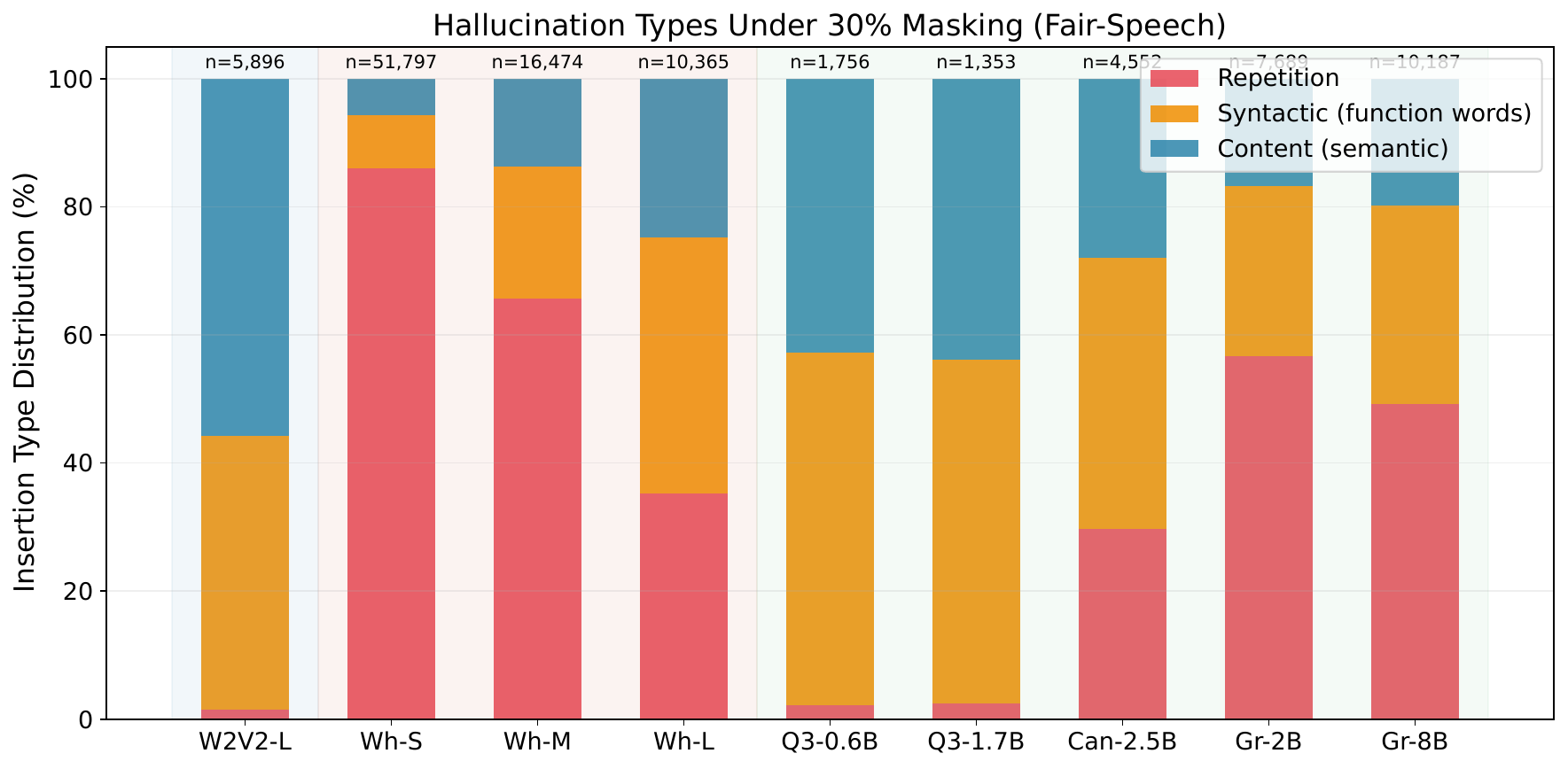}
\caption{Hallucination type distribution under masking. $n$ denotes insertion count. Whisper-small is dominated by repetition loops; Qwen3 shows near-zero repetition; Granite reintroduces repetition due to high compression.}
\label{fig:hallucination_masking_fs}
\end{figure}

\paragraph{Compression reintroduces repetition.}
Among Gen~3 models, audio compression predicts hallucination type under degradation (Figure~\ref{fig:hallucination_masking_fs}).
Qwen3 (low compression) maintains $<$4\% repetition across all conditions.
But Granite-Speech (high compression, Q-former bottleneck) exhibits 49-57\% repetition under 30\% masking, closer to Whisper than to Qwen, with 7{,}689 (2B) and 10{,}187 (8B) total insertions.
Canary (medium compression) falls between at 30\% repetition and 4{,}552 insertions.
Just as compression predicts accent fairness on clean audio (\S\ref{sec:accent:compression}), it also predicts hallucination behavior under degradation: high audio compression induces Whisper-like repetition pathology even when the decoder is an explicit LLM.
Silence perturbations show similar patterns: Whisper-small produces 57-60\% repetition, Qwen3 $<$4\%, and Granite 31-43\% (full breakdowns in Appendix~\ref{app:perturbation_full}).

\section{Discussion}
\label{sec:discussion}

\subsection{Accuracy does not imply fairness; degradation does not imply inequity}
\label{sec:discussion:pareto}

\begin{figure}[t]
\centering
\begin{subfigure}[t]{0.49\textwidth}
  \centering
  \includegraphics[width=\textwidth]{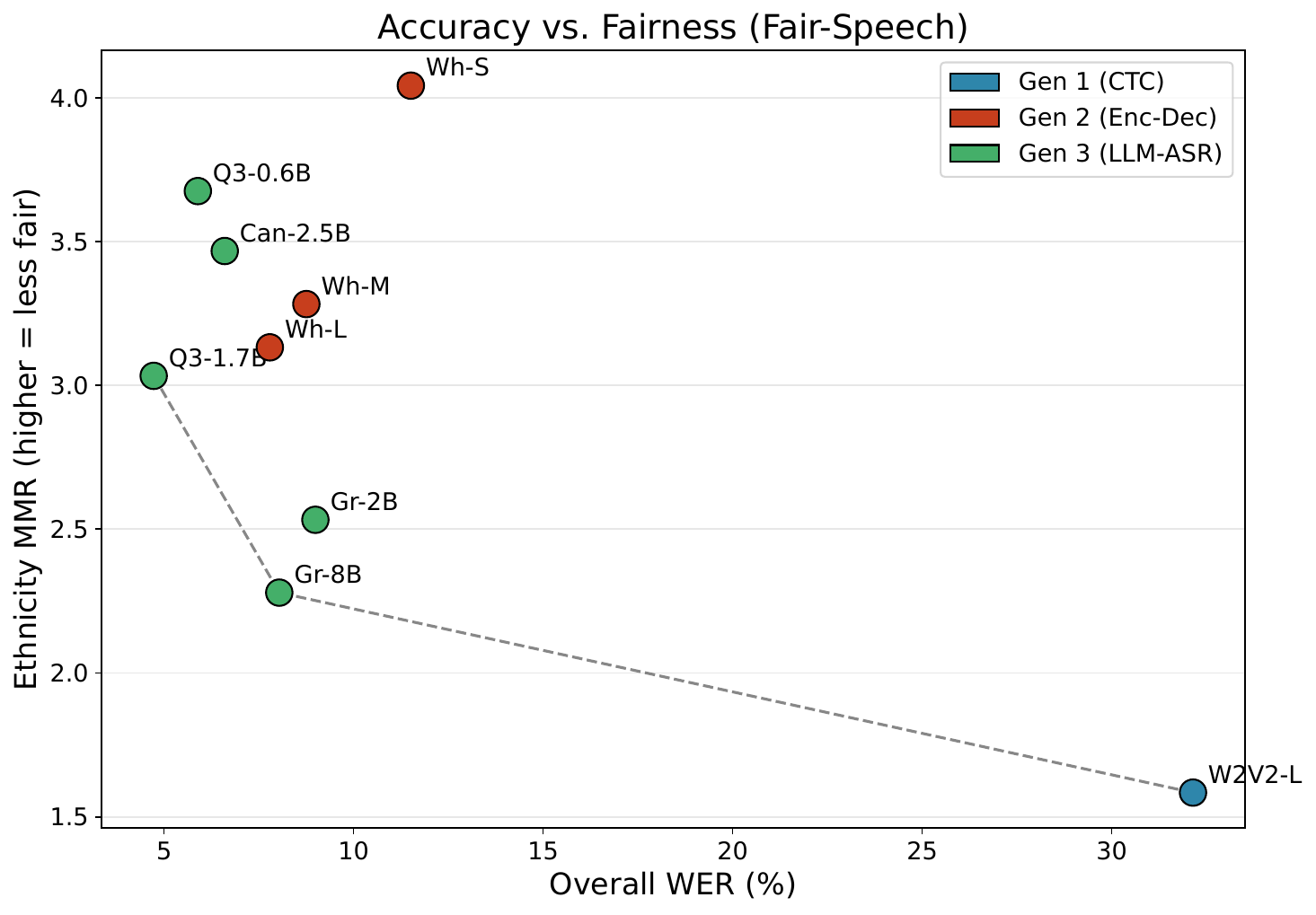}
  \caption{Clean audio.}
  \label{fig:pareto_clean}
\end{subfigure}
\hfill
\begin{subfigure}[t]{0.49\textwidth}
  \centering
  \includegraphics[width=\textwidth]{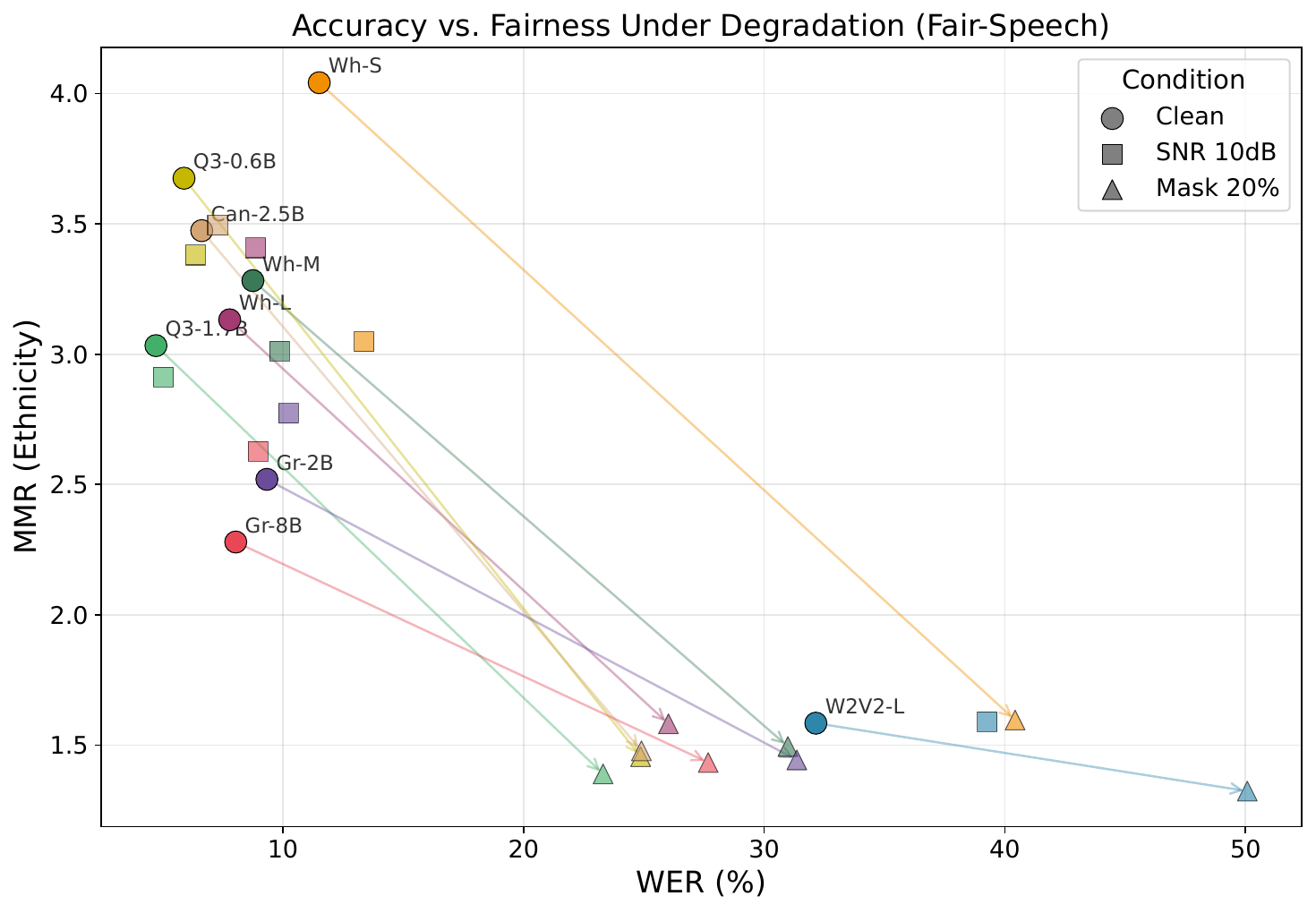}
  \caption{Under degradation (SNR 10\,dB and 20\% masking).}
  \label{fig:pareto_perturbed}
\end{subfigure}
\caption{Accuracy vs.\ ethnicity fairness. (a) Clean: Qwen3-1.7B and Granite-8B define Pareto frontier. (b) Under degradation: models converge to degenerate frontier where low disparity coexists with poor performance. Arrows trace clean to 20\% masking.}
\label{fig:pareto}
\end{figure}

Figure~\ref{fig:pareto} synthesizes the accuracy-fairness tradeoff.
On clean speech (panel a), Qwen3-1.7B achieves the best WER (4.73\%) with competitive ethnicity fairness (MMR\,{=}\,3.03), while Granite-8B achieves the best ethnicity fairness among competitive models (MMR\,{=}\,2.28) at higher WER (8.04\%).
On the accent axis, Qwen3-1.7B also achieves the best MMR (1.82), making it the strongest overall candidate for fair deployment.
Under degradation (panel b), all models shift rightward (higher WER) and downward (lower MMR), converging toward a degenerate frontier where low relative disparity coexists with poor absolute performance, echoing the ``low-accuracy parity'' paradox from \S\ref{sec:ethnicity:paradox}.

\noindent\textbf{Practical recommendations.}
Qwen3-1.7B dominates on accuracy, accent fairness, and robustness; Granite-8B is preferable for ethnicity fairness.
Whisper-large-v3 should not be deployed on accented speech without hallucination mitigation \citep{baranski2025whisper}, especially with silence or signal dropout.
Low-compression audio encoding (as in Qwen3) provides robustness against both accuracy loss and bias amplification.

\subsection{Limitations}
\label{sec:discussion:limitations}

Our study is limited to English read and prompted speech and may not generalize to multilingual or spontaneous ASR.
Training data is a confound: we cannot fully disentangle the decoder's contribution from training corpora differences.
Granite-8B's higher LibriSpeech WER than Granite-2B is likely a verbose generation artifact and does not affect fairness comparisons.
Perturbation conditions are synthetic and applied in isolation.
Common Voice sentences are drawn from Wikipedia, and LibriSpeech from Project Gutenberg, both standard components of LLM pretraining corpora.
Gen~3 models' LLM decoders may have encountered exact transcripts during pretraining.
Two observations constrain this concern: if contamination drove improvements, we would expect uniform WER gains across groups, yet Black/AA gaps persist (+203\% for Qwen3-1.7B); and Gen 3 models maintain their advantage even under severe masking that removes intelligible acoustic content. In such conditions, memorized text cannot aid decoding, suggesting that the observed gains are unlikely to be driven by test-set contamination

\vspace{-1ex}
\section{Related work}
\label{sec:related}
\vspace{-0.5ex}
ASR demographic bias is extensively documented across ethnicity \citep{koenecke2020racial}, gender and dialect \citep{tatman2017gender}, and Indian-accented English \citep{javed2023svarah}. While recent multi-axis benchmarks and controlled datasets like Fair-Speech offer standardized evaluation \citep{rai2025fairbench, veliche2024fairspeech}, they primarily assess commercial systems. Unlike these studies, which evaluate limited systems on isolated axes, we specifically investigate how decoder architecture and language model integration drive these disparities. On hallucination, \citet{koenecke2024careless} found ${\sim}$1\% of Whisper outputs hallucinate (38\% harmful); \citet{baranski2025whisper} showed non-speech audio triggers hallucinations and proposed energy-based filtering; \citet{frieske2024hallucinations} developed taxonomy-based detection; and \citet{atwany2025} traced hallucinations to distribution shifts. \citep{shallow2025} introduced a hallucination benchmark with a fine-grained taxonomy across architectures and acoustic conditions, but none of these studies connect hallucination to demographic bias.  Robustness studies further highlight performance drops under noise and domain shifts \citep{shah2025robustbench, slam2025}, yet demographic breakdowns in these contexts remain sparse. \citet{wei-etal-2026-bias} examine bias in spoken QA for multimodal LLMs but do not study transcription quality or decoder architecture. No prior study examines whether adding a pretrained LLM decoder systematically affects fairness under acoustic degradation.

\section{Conclusion}
\label{sec:conclusion}
\vspace{-0.5ex}
We studied how three generations of language model integration affect demographic fairness across nine models, five demographic axes, and 12 degradation conditions.
LLM decoders do not amplify racial bias, but stronger priors inflate relative disparity as majority-group WER approaches zero.
Severe degradation generally compresses fairness gaps, with one critical exception: silence amplifies Whisper's accent bias ($\alpha$\,{=}\,4.64) via demographic-selective hallucination.
Under masking, Whisper produces catastrophic repetition loops (86\% of insertions) while explicit-LLM decoders produce 38$\times$ fewer with near-zero repetition; high audio compression reintroduces this pathology, identifying audio encoder architecture as the primary lever for equitable recognition.

\vspace{-0.5ex}
\section*{Reproducibility Statement}
\vspace{-0.5ex}
To facilitate reproducibility, all model weights and their HuggingFace identifiers, alongside our exact inference parameters and text normalization steps, are fully reported in Appendix~\ref{app:inference}. The Fair-Speech, Common Voice~24, MUSAN, and OpenSLR RIR datasets are publicly available. All code for our data preprocessing, perturbation generation, and evaluation pipeline will be open-sourced upon publication.

\vspace{-0.5ex}
\section*{Ethics Statement}
\vspace{-0.5ex}
This study evaluates ASR systems for demographic bias to inform fairer deployment.
All demographic labels are self-reported; we report group-level statistics only and no individual speakers can be identified.
Our finding that Black/African-American speakers experience the highest WER across all nine models should motivate targeted improvement, not rationalize exclusion from ASR-dependent services.
Hallucination examples (Table~\ref{tab:hallucination_examples}) illustrate failure modes from public datasets.
Our fairness metrics are imperfect proxies for harm; a low MMR can mask uniformly poor performance (\S\ref{sec:ethnicity:paradox}).
Practitioners should evaluate deployment-specific impacts beyond aggregate metrics.

\bibliography{references}
\bibliographystyle{colm2026_conference}

\clearpage
\appendix

\section{Ethnicity and Accent WER tables}
\label{app:ethnicity}

Table~\ref{tab:fs_wer_ethnicity} reports WER by ethnicity on Fair-Speech; Table~\ref{tab:wer_accent} reports WER by accent on Common Voice~24.
\label{app:accent}

\begin{table}[t]
\centering
\caption{WER (\%) by ethnicity across ASR models on Fair-Speech.}
\label{tab:fs_wer_ethnicity}
\resizebox{\textwidth}{!}{
\begin{tabular}{lrrrrrrrrr}
\toprule
Model & White & Black/AA & Hispanic & Asian & Native American & Pacific Islander & Middle Eastern & MMR & Gap (\%) \\
\midrule
wav2vec2-large & 26.3 & 41.6 & 30.2 & 28.7 & 27.3 & 30.4 & 32.5 & 1.58 & 58.4 \\
whisper-small & 10.5 & 18.6 & 7.6 & 4.6 & 9.7 & 6.4 & 9.5 & 4.04 & 304.2 \\
whisper-medium & 9.8 & 12.4 & 6.2 & 3.8 & 7.5 & 4.1 & 7.6 & 3.28 & 228.2 \\
whisper-large-v3 & 9.4 & 10.2 & 5.6 & 3.3 & 7.4 & 3.7 & 6.1 & 3.13 & 213.2 \\
qwen3-asr-0.6b & 3.0 & 11.0 & 3.8 & 3.5 & 4.1 & 4.4 & 5.8 & 3.68 & 267.5 \\
qwen3-asr-1.7b & 2.8 & 8.5 & 2.9 & 2.8 & 3.6 & 3.5 & 4.4 & 3.03 & 203.3 \\
canary-qwen-2.5b & 4.9 & 11.7 & 3.8 & 3.4 & 4.7 & 4.7 & 5.8 & 3.47 & 246.7 \\
granite-speech-3.3-2b & 10.0 & 13.4 & 5.6 & 5.3 & 6.2 & 5.4 & 6.3 & 2.53 & 153.3 \\
granite-speech-3.3-8b & 7.6 & 12.1 & 5.5 & 6.1 & 5.6 & 5.3 & 5.6 & 2.28 & 127.9 \\
\bottomrule
\end{tabular}
}
\end{table}
\begin{table}[t]
\centering
\caption{WER (\%) by accent across ASR models on Common Voice~24.}
\label{tab:wer_accent}
\resizebox{\textwidth}{!}{
\begin{tabular}{lrrrrrrrr}
\toprule
Model & African & Australia & Canada & England & Indian & Us & MMR & Gap (\%) \\
\midrule
wav2vec2-large & 23.3 & 13.2 & 11.8 & 16.5 & 25.2 & 16.4 & 2.13 & 113.0 \\
whisper-small & 20.3 & 8.0 & 6.3 & 13.1 & 17.6 & 11.5 & 3.23 & 222.9 \\
whisper-medium & 11.5 & 6.4 & 4.2 & 9.4 & 13.2 & 8.8 & 3.15 & 215.3 \\
whisper-large-v3 & 13.0 & 5.2 & 3.6 & 7.7 & 19.0 & 7.5 & 5.34 & 434.0 \\
qwen3-asr-0.6b & 9.6 & 5.3 & 4.1 & 8.1 & 11.1 & 7.5 & 2.72 & 171.6 \\
qwen3-asr-1.7b & 8.0 & 4.4 & 4.6 & 6.7 & 7.8 & 5.8 & 1.82 & 82.0 \\
canary-qwen-2.5b & 8.0 & 4.0 & 4.8 & 6.3 & 7.2 & 5.8 & 2.00 & 100.2 \\
granite-speech-3.3-2b & 10.0 & 5.0 & 3.1 & 9.0 & 10.4 & 8.3 & 3.30 & 230.2 \\
granite-speech-3.3-8b & 16.1 & 4.7 & 4.3 & 9.4 & 12.9 & 10.6 & 3.74 & 274.5 \\
\bottomrule
\end{tabular}
}
\end{table}

\section{Whisper scaling trajectory}
\label{app:whisper_scaling}

\begin{figure}[ht]
\centering
\includegraphics[width=\textwidth]{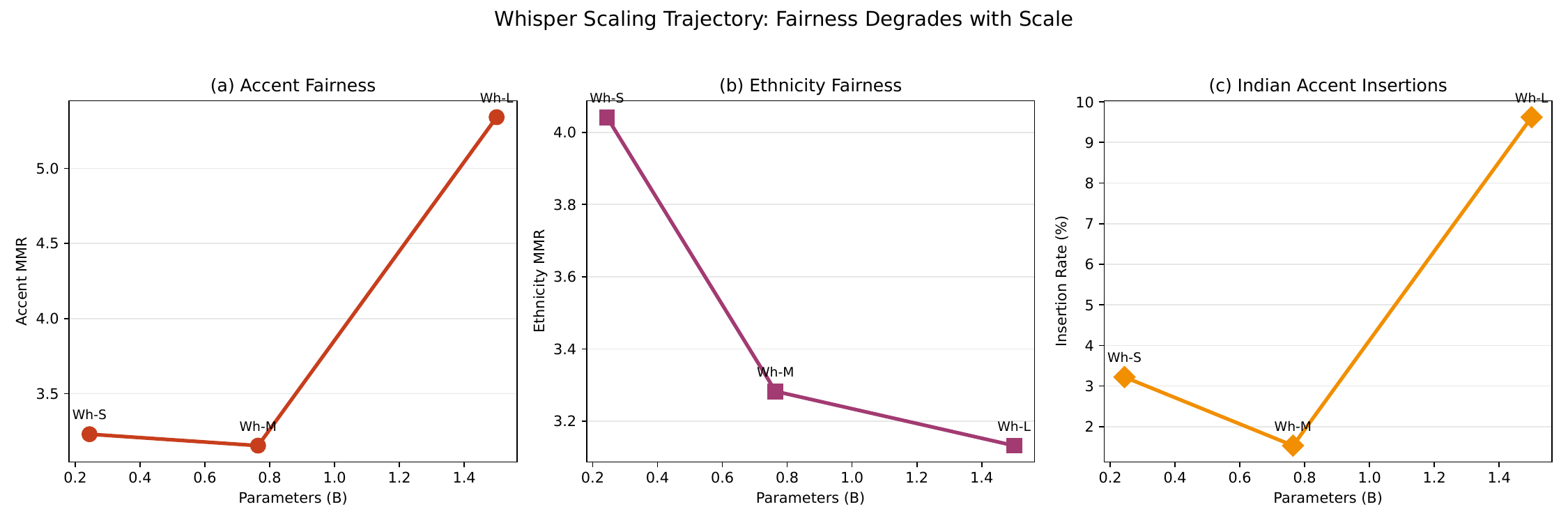}
\caption{Whisper three-point scaling trajectory.
  Accent MMR worsens at large-v3 (5.34) even as ethnicity MMR improves (3.13), driven by the Indian-accent insertion pathology.}
\label{fig:whisper_scaling}
\end{figure}

Figure~\ref{fig:whisper_scaling} shows the Whisper three-point scaling trajectory.
Accent MMR worsens at large-v3 even as ethnicity MMR improves, driven by the Indian-accent insertion pathology (\S\ref{sec:accent:hallucination}).
Scaling within Whisper shows architecture-dependent effects: ethnicity fairness improves (MMR: 4.04$\to$3.28$\to$3.13) while accent fairness worsens (MMR: 3.23$\to$3.15$\to$5.34).

\section{Gender WER}
\label{app:gender}

\begin{figure}[h]
\centering
\includegraphics[width=\textwidth]{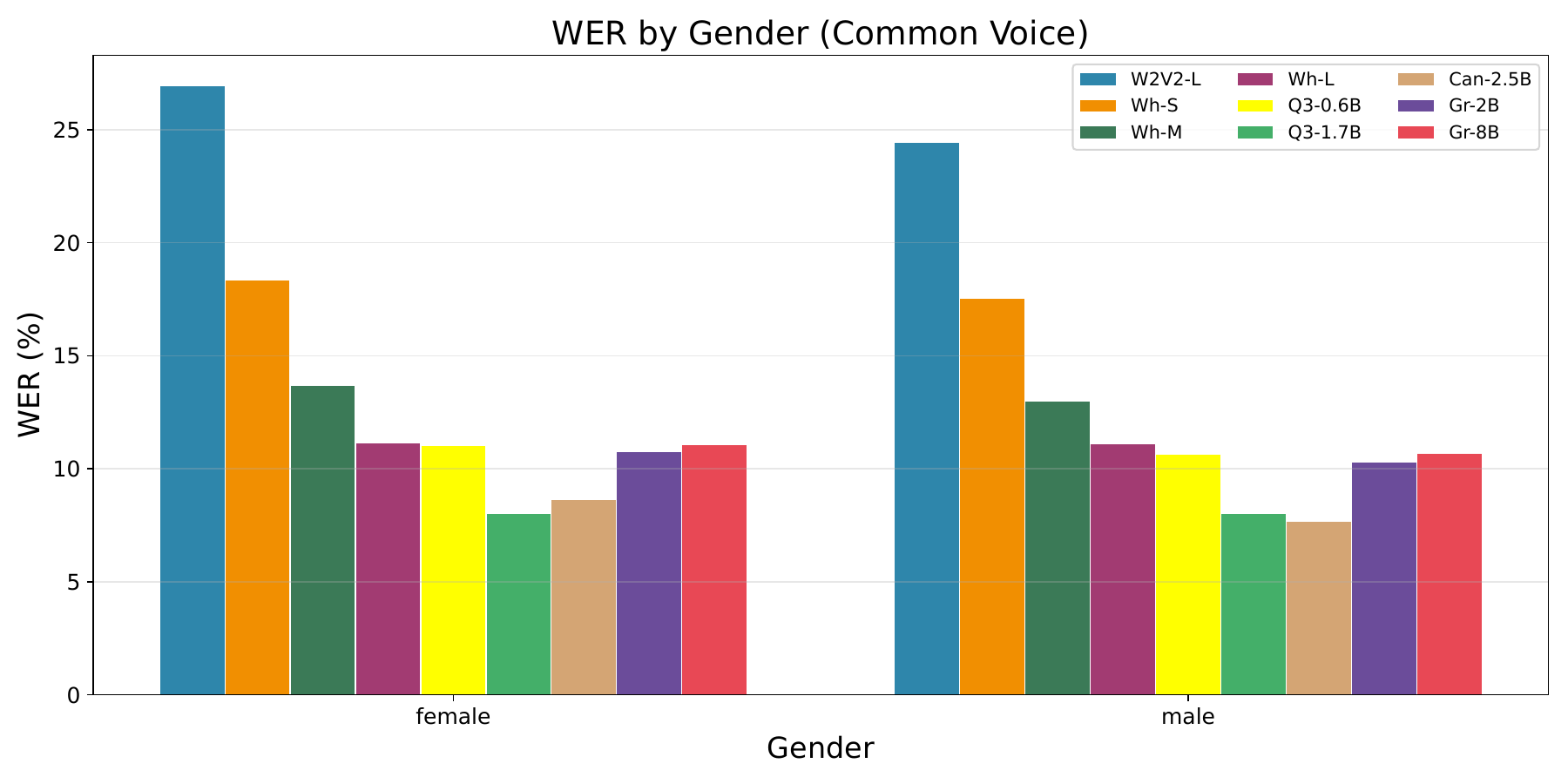}
\caption{WER by gender on Common Voice~24.  Gender gaps are minimal across all models (max MMR\,{=}\,1.12).}
\label{fig:gender_cv}
\end{figure}

\begin{figure}[h]
\centering
\includegraphics[width=\textwidth]{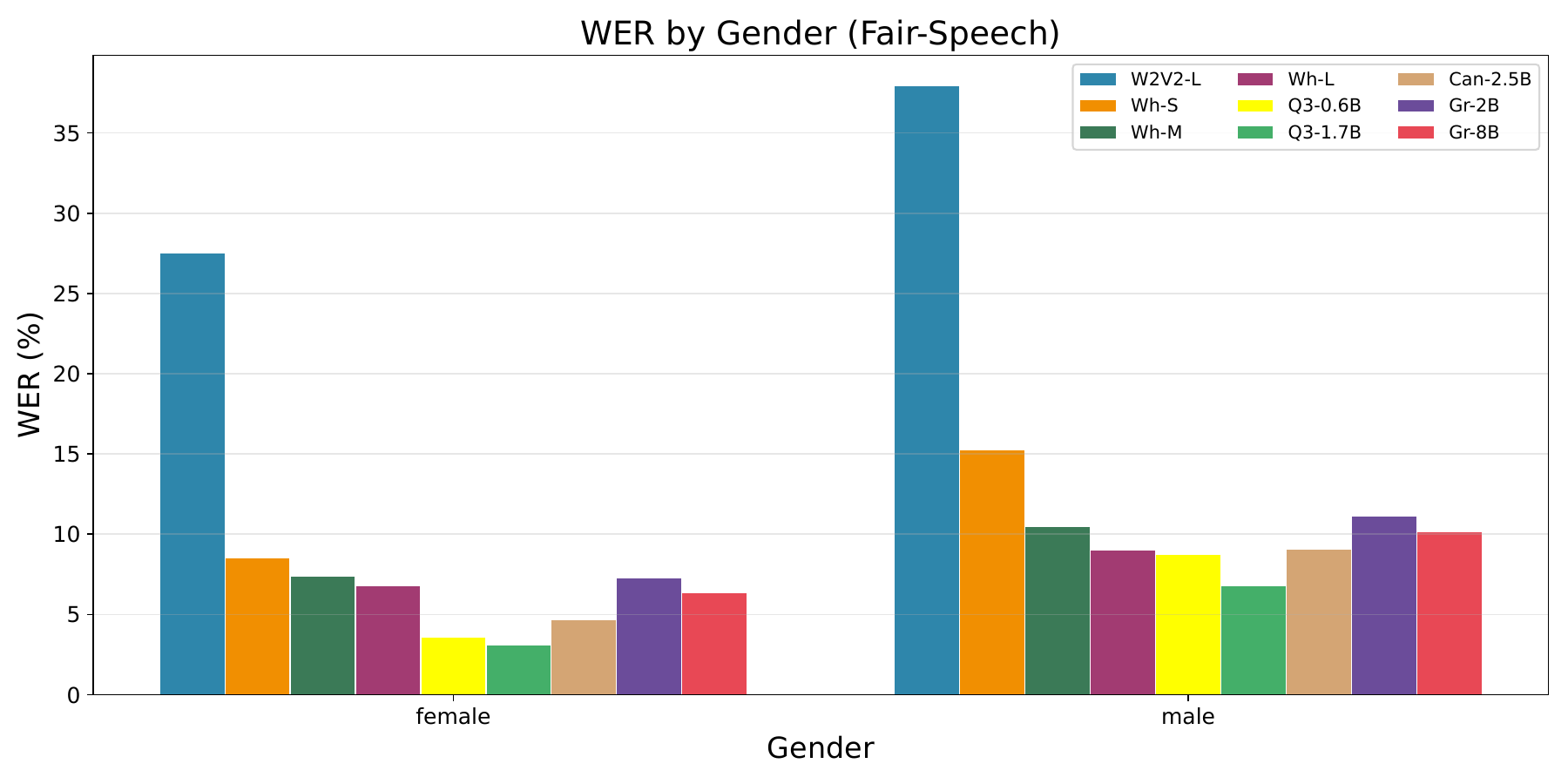}
\caption{WER by gender on Fair-Speech.  Males are consistently 1.5-2.4$\times$ worse than females across Gen~3 models, revealed by Fair-Speech's controlled prompts.}
\label{fig:gender_fs}
\end{figure}

Table~\ref{tab:wer_gender} reports WER by gender on Common Voice~24 and Table~\ref{tab:fs_wer_gender} on Fair-Speech.
As noted in \S\ref{sec:ethnicity}, gender disparities are negligible on Common Voice (max MMR\,{=}\,1.12) but substantial on Fair-Speech (Qwen3-0.6B gender MMR\,{=}\,2.42).
Males are consistently 1.5-2.4$\times$ worse than females across Gen~3 models on Fair-Speech.
This discrepancy likely reflects Fair-Speech's controlled prompts, which reveal acoustic-level gender effects masked by Common Voice's heterogeneous sentences.
Figures~\ref{fig:gender_cv} and~\ref{fig:gender_fs} visualize these differences.
\begin{table}[htp]
\centering
\begin{minipage}[t]{0.48\textwidth}
\caption{WER (\%) by gender across ASR models on Common Voice~24.}
\label{tab:wer_gender}
\resizebox{\linewidth}{!}{
\begin{tabular}{lrrrr}
\toprule
Model & Female & Male & MMR & Gap (\%) \\
\midrule
wav2vec2-large & 26.9 & 24.5 & 1.10 & 10.2 \\
whisper-small & 18.3 & 17.5 & 1.05 & 4.6 \\
whisper-medium & 13.7 & 13.0 & 1.06 & 5.6 \\
whisper-large-v3 & 11.1 & 11.1 & 1.00 & 0.2 \\
qwen3-asr-0.6b & 11.0 & 10.6 & 1.04 & 3.8 \\
qwen3-asr-1.7b & 8.0 & 8.0 & 1.00 & 0.2 \\
canary-qwen-2.5b & 8.6 & 7.7 & 1.12 & 12.3 \\
granite-speech-3.3-2b & 10.7 & 10.3 & 1.04 & 4.5 \\
granite-speech-3.3-8b & 11.0 & 10.7 & 1.03 & 3.3 \\
\bottomrule
\end{tabular}
}
\end{minipage}
\hfill
\begin{minipage}[t]{0.48\textwidth}
\caption{WER (\%) by gender across ASR models on Fair-Speech.}
\label{tab:fs_wer_gender}
\resizebox{\linewidth}{!}{
\begin{tabular}{lrrrr}
\toprule
Model & female & male & MMR & Gap (\%) \\
\midrule
wav2vec2-large & 27.5 & 37.9 & 1.38 & 37.7 \\
whisper-small & 8.5 & 15.3 & 1.79 & 79.5 \\
whisper-medium & 7.4 & 10.5 & 1.42 & 42.4 \\
whisper-large-v3 & 6.8 & 9.0 & 1.33 & 32.8 \\
qwen3-asr-0.6b & 3.6 & 8.8 & 2.42 & 142.4 \\
qwen3-asr-1.7b & 3.1 & 6.8 & 2.18 & 118.0 \\
canary-qwen-2.5b & 4.6 & 9.1 & 1.95 & 95.0 \\
granite-speech-3.3-2b & 7.3 & 11.2 & 1.53 & 53.3 \\
granite-speech-3.3-8b & 6.4 & 10.1 & 1.59 & 59.1 \\
\bottomrule
\end{tabular}
}
\end{minipage}
\end{table}

\section{Age WER}
\label{app:age}

\begin{figure}[h]
\centering
\includegraphics[width=\textwidth]{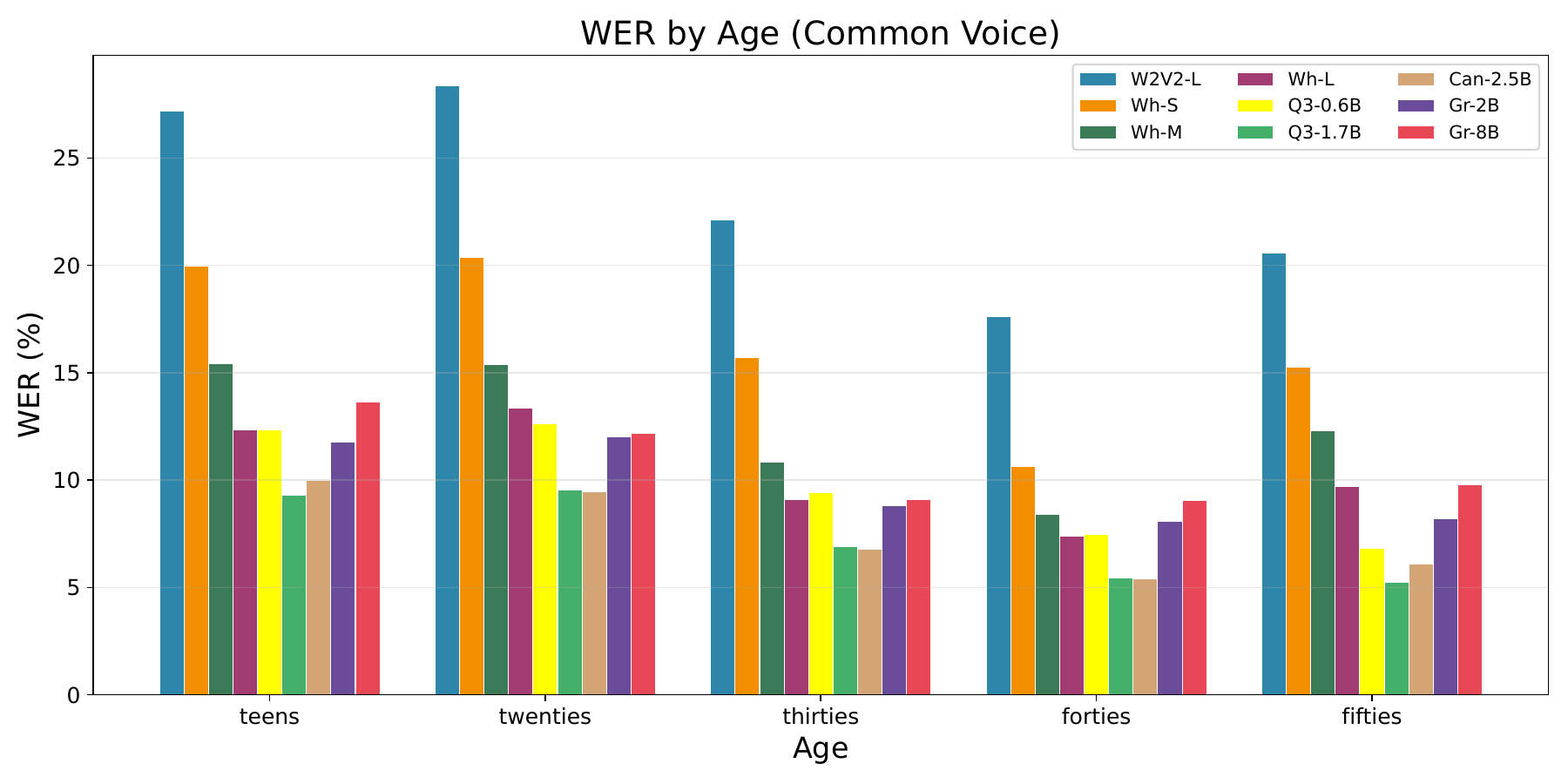}
\caption{WER by age group on Common Voice~24.}
\label{fig:age_cv}
\end{figure}

\begin{figure}[h]
\centering
\includegraphics[width=\textwidth]{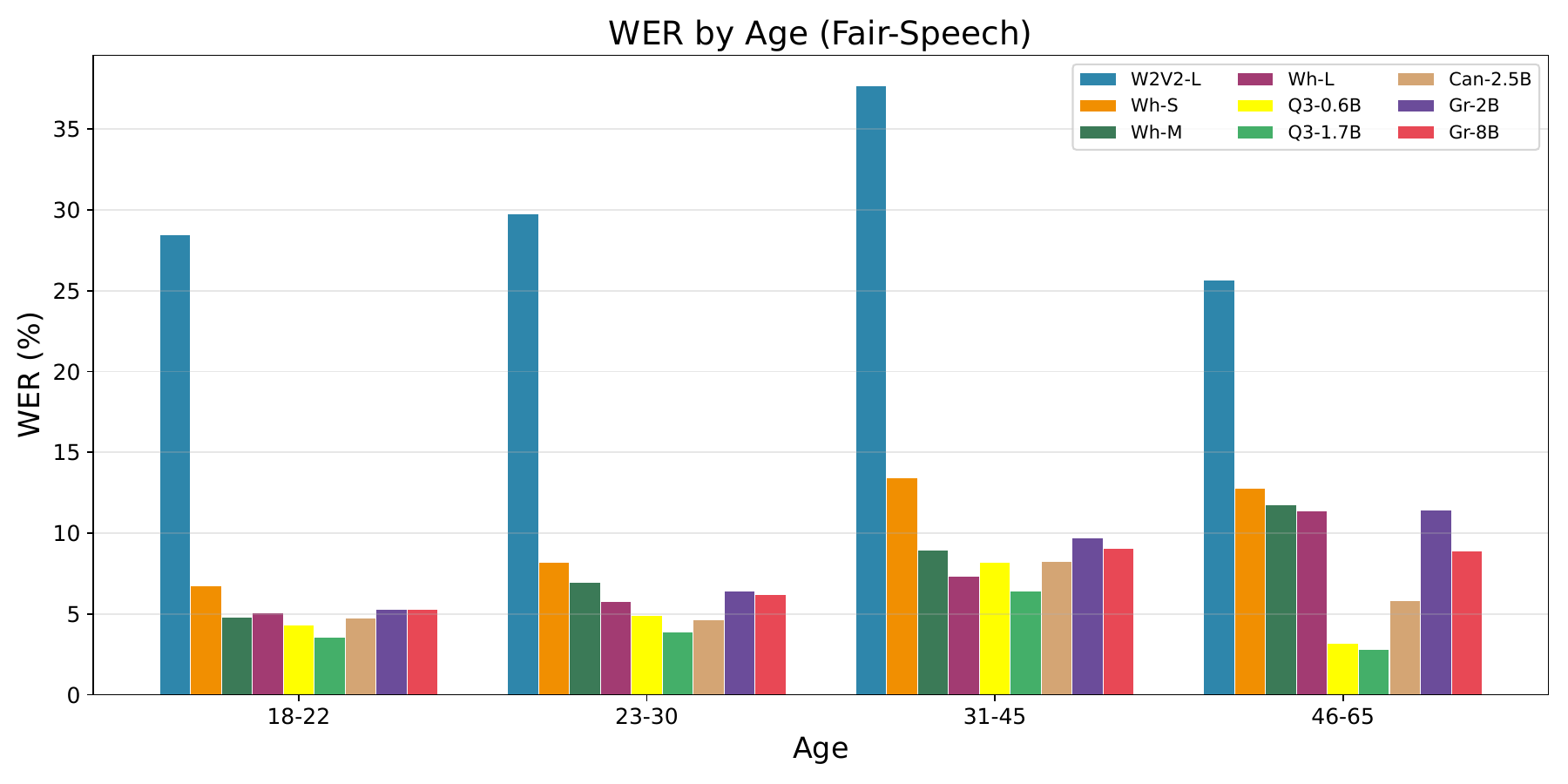}
\caption{WER by age group on Fair-Speech.  Note the age$\times$architecture interaction: the 46-65 group flips from worst (Whisper) to best (Qwen3).}
\label{fig:age_fs}
\end{figure}

Tables~\ref{tab:wer_age} and~\ref{tab:fs_wer_age} report WER by age group on Common Voice~24 and Fair-Speech, respectively.
A notable age$\times$architecture interaction emerges on Fair-Speech: the 46-65 age group is the \emph{worst} for Whisper models (Whisper-medium: 11.8\% WER) but the \emph{best} for Qwen3 models (Qwen3-1.7B: 2.8\%).
Figure~\ref{fig:age_cv} shows the Common Voice results; Figure~\ref{fig:age_fs} shows Fair-Speech.

\begin{table}[t]
\centering
\caption{WER (\%) by age across ASR models on Common Voice~24.}
\label{tab:wer_age}
\resizebox{\textwidth}{!}{
\begin{tabular}{lrrrrrrr}
\toprule
Model & Teens & Twenties & Thirties & Forties & Fifties & MMR & Gap (\%) \\
\midrule
wav2vec2-large & 27.2 & 28.4 & 22.1 & 17.6 & 20.6 & 1.61 & 61.1 \\
whisper-small & 20.0 & 20.4 & 15.7 & 10.6 & 15.3 & 1.92 & 92.0 \\
whisper-medium & 15.4 & 15.4 & 10.8 & 8.4 & 12.3 & 1.84 & 84.1 \\
whisper-large-v3 & 12.3 & 13.4 & 9.1 & 7.4 & 9.7 & 1.81 & 81.2 \\
qwen3-asr-0.6b & 12.3 & 12.6 & 9.4 & 7.5 & 6.8 & 1.85 & 85.1 \\
qwen3-asr-1.7b & 9.3 & 9.5 & 6.9 & 5.4 & 5.2 & 1.82 & 82.5 \\
canary-qwen-2.5b & 10.0 & 9.4 & 6.8 & 5.4 & 6.1 & 1.85 & 85.4 \\
granite-speech-3.3-2b & 11.8 & 12.0 & 8.8 & 8.1 & 8.2 & 1.48 & 48.4 \\
granite-speech-3.3-8b & 13.7 & 12.2 & 9.1 & 9.1 & 9.8 & 1.51 & 50.8 \\
\bottomrule
\end{tabular}
}
\end{table}

\begin{table}[t]
\centering
\caption{WER (\%) by age across ASR models on Fair-Speech.}
\label{tab:fs_wer_age}
\begin{tabular}{lrrrrrr}
\toprule
Model & 18-22 & 23-30 & 31-45 & 46-65 & MMR & Gap (\%) \\
\midrule
wav2vec2-large & 28.5 & 29.7 & 37.7 & 25.7 & 1.47 & 46.9 \\
whisper-small & 6.8 & 8.2 & 13.4 & 12.8 & 1.98 & 98.3 \\
whisper-medium & 4.8 & 7.0 & 8.9 & 11.8 & 2.44 & 144.3 \\
whisper-large-v3 & 5.0 & 5.8 & 7.4 & 11.4 & 2.25 & 125.1 \\
qwen3-asr-0.6b & 4.3 & 4.9 & 8.2 & 3.2 & 2.60 & 160.1 \\
qwen3-asr-1.7b & 3.5 & 3.9 & 6.4 & 2.8 & 2.30 & 130.5 \\
canary-qwen-2.5b & 4.8 & 4.7 & 8.3 & 5.8 & 1.77 & 76.9 \\
granite-speech-3.3-2b & 5.3 & 6.4 & 9.7 & 11.4 & 2.15 & 115.2 \\
granite-speech-3.3-8b & 5.3 & 6.2 & 9.0 & 8.9 & 1.71 & 70.7 \\
\bottomrule
\end{tabular}
\end{table}

\section{Socioeconomic status WER}
\label{app:ses}

\begin{figure}[h]
\centering
\includegraphics[width=\textwidth]{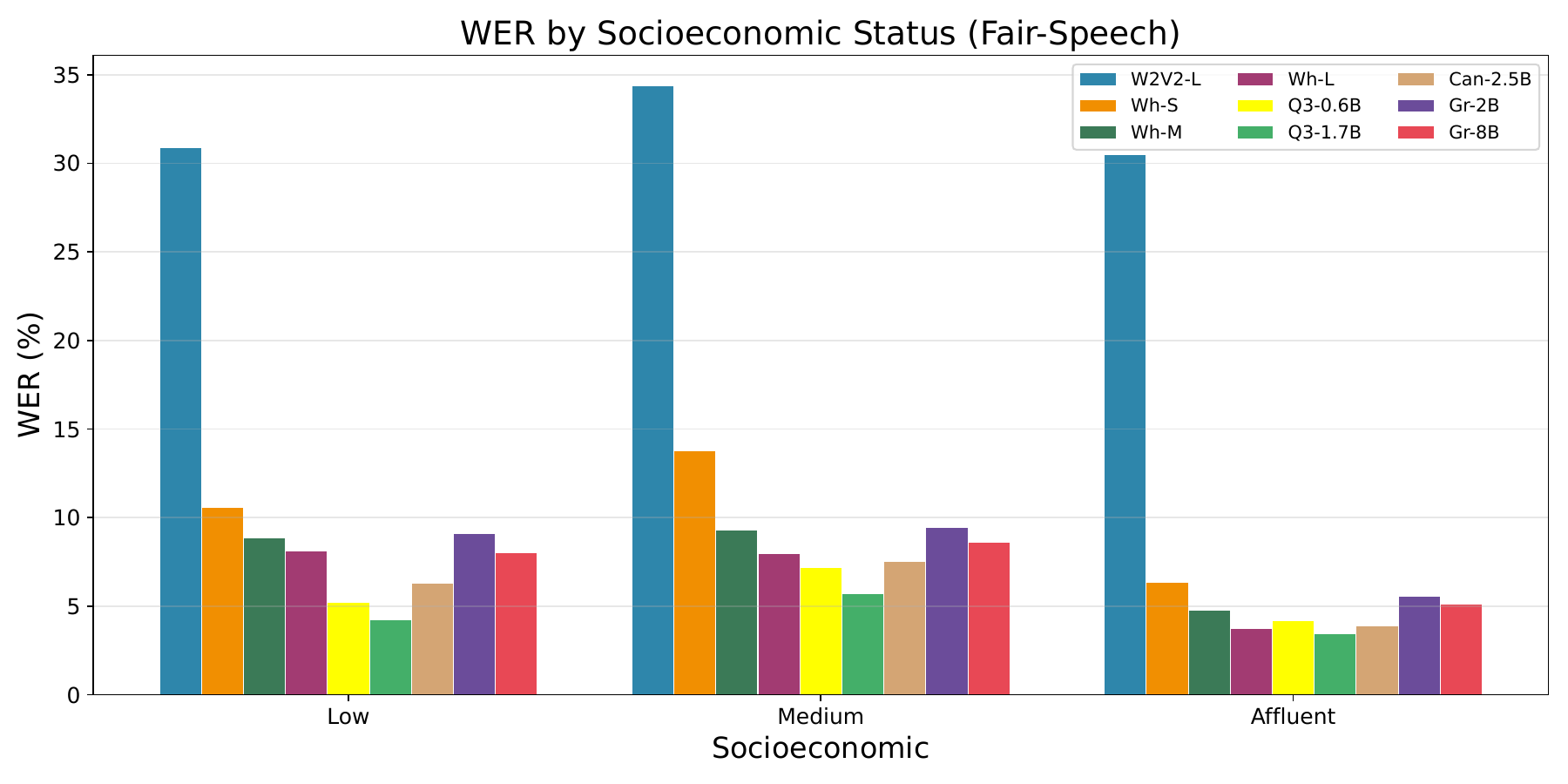}
\caption{WER by socioeconomic status on Fair-Speech.  Medium-SES speakers have the highest WER across most models, with Affluent consistently lowest.}
\label{fig:ses_fs}
\end{figure}

Table~\ref{tab:fs_wer_socioeconomic} reports WER by self-reported socioeconomic status on Fair-Speech.
Medium-SES speakers have the highest WER across 8 of 9 models (e.g., Whisper-small: Medium 13.8\% vs.\ Low 10.6\% vs.\ Affluent 6.4\%), with Affluent consistently lowest (SES MMR: 1.13-2.17).
Gen~3 models show slightly narrower gaps (MMR 1.67-1.93) than Gen~2 (1.94-2.17).
Figure~\ref{fig:ses_fs} visualizes the gap.

\begin{table}[t]
\centering
\caption{WER (\%) by socioeconomic status across ASR models on Fair-Speech.}
\label{tab:fs_wer_socioeconomic}
\begin{tabular}{lrrrrr}
\toprule
Model & Low & Medium & Affluent & MMR & Gap (\%) \\
\midrule
wav2vec2-large & 30.9 & 34.4 & 30.5 & 1.13 & 12.8 \\
whisper-small & 10.6 & 13.8 & 6.4 & 2.17 & 116.7 \\
whisper-medium & 8.8 & 9.3 & 4.8 & 1.94 & 94.3 \\
whisper-large-v3 & 8.1 & 8.0 & 3.7 & 2.17 & 117.0 \\
qwen3-asr-0.6b & 5.2 & 7.2 & 4.2 & 1.72 & 72.2 \\
qwen3-asr-1.7b & 4.2 & 5.7 & 3.4 & 1.67 & 67.0 \\
canary-qwen-2.5b & 6.3 & 7.5 & 3.9 & 1.93 & 92.5 \\
granite-speech-3.3-2b & 9.1 & 9.5 & 5.5 & 1.70 & 70.5 \\
granite-speech-3.3-8b & 8.0 & 8.6 & 5.1 & 1.68 & 68.3 \\
\bottomrule
\end{tabular}
\end{table}

\section{Full first-language WER (21 languages)}
\label{app:l1}

\begin{figure}[ht]
\centering
\includegraphics[width=\textwidth]{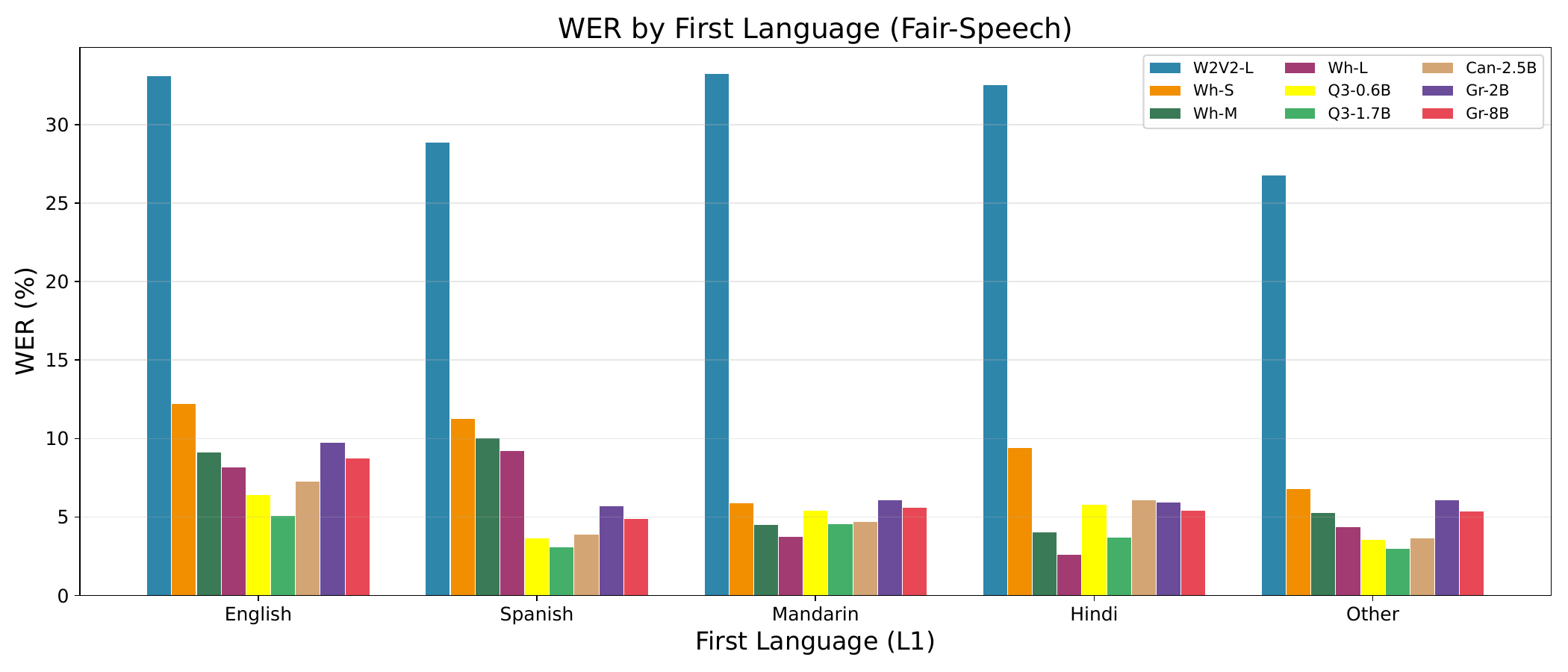}
\caption{WER by first-language group on Fair-Speech.
  L1-English speakers have the highest WER across most models (an ethnicity confound).
  Whisper models show extreme L1 sensitivity (MMR up to 15.05 across individual L1s), while Gen~3 models are substantially more robust (MMR\,$\leq$\,5.38).}
\label{fig:l1_analysis}
\end{figure}

Whisper's implicit-LM decoder creates extreme first-language (L1) sensitivity on Fair-Speech.
Across 21 individual L1s (Table~\ref{tab:fs_wer_first_language_land}), Whisper-small has an L1 MMR of 15.05 (Filipino: 1.0\%, French: 14.3\%).
Whisper-large-v3 shows a 2.2$\times$ ratio between L1-English (8.2\%) and L1-Mandarin (3.8\%) speakers.
Gen~3 models are substantially more robust: Qwen3-1.7B narrows this to only 1.1$\times$ (L1-English 5.1\% vs.\ L1-Mandarin 4.6\%).
The counterintuitive finding that L1-English speakers have the highest WER is an ethnicity confound: L1-English speakers in Fair-Speech are disproportionately Black/AA and Native American.
Figure~\ref{fig:l1_analysis} visualizes results by L1 group.

\begin{landscape}
\centering
\captionof{table}{WER (\%) by individual first language (21 L1s with $n \geq 50$) across ASR models on Fair-Speech.}
\label{tab:fs_wer_first_language_land}
\vspace{1em}
\resizebox{0.95\linewidth}{!}{
\begin{tabular}{lrrrrrrrrrrrrrrrrrrrrrrr}
\toprule
Model & Arabic & Cantonese & Dutch & English & Filipino & French & German & Hindi & Indonesian & Korean & Malay & Mandarin & Marathi & Nepali & Other & Portuguese & Russian & Spanish & Tagalog & Urdu & Vietnamese & MMR & Gap (\%) \\
\midrule
wav2vec2-large & 29.6 & 33.4 & 40.8 & 33.1 & 23.3 & 20.7 & 29.8 & 32.6 & 26.0 & 30.2 & 25.5 & 33.3 & 32.7 & 27.1 & 24.4 & 29.6 & 39.0 & 28.9 & 23.7 & 21.6 & 27.7 & 1.97 & 96.7 \\
whisper-small & 4.7 & 5.6 & 4.6 & 12.2 & 1.0 & 14.3 & 10.4 & 9.4 & 4.9 & 2.7 & 2.9 & 5.9 & 4.2 & 2.5 & 9.1 & 4.9 & 5.7 & 11.3 & 3.3 & 5.6 & 4.7 & 15.05 & 1404.9 \\
whisper-medium & 4.0 & 4.0 & 3.5 & 9.2 & 2.4 & 8.9 & 8.1 & 4.0 & 5.9 & 2.0 & 3.8 & 4.5 & 1.0 & 1.7 & 6.8 & 3.1 & 4.9 & 10.0 & 2.0 & 5.1 & 4.7 & 9.67 & 866.5 \\
whisper-large-v3 & 3.3 & 3.2 & 2.3 & 8.2 & 1.0 & 7.1 & 3.4 & 2.6 & 4.9 & 1.5 & 2.8 & 3.8 & 2.1 & 1.7 & 6.3 & 3.2 & 3.6 & 9.2 & 2.2 & 4.4 & 3.9 & 9.72 & 871.9 \\
qwen3-asr-0.6b & 3.8 & 3.2 & 4.2 & 6.4 & 1.0 & 4.3 & 2.7 & 5.8 & 5.2 & 2.6 & 1.1 & 5.4 & 2.9 & 1.9 & 3.9 & 3.1 & 6.1 & 3.7 & 2.2 & 3.4 & 3.4 & 6.74 & 574.3 \\
qwen3-asr-1.7b & 3.3 & 2.6 & 3.5 & 5.1 & 1.0 & 3.1 & 2.3 & 3.7 & 4.9 & 1.5 & 1.4 & 4.6 & 1.8 & 1.9 & 3.3 & 3.5 & 3.7 & 3.1 & 1.9 & 2.3 & 3.8 & 5.38 & 437.9 \\
canary-qwen-2.5b & 4.4 & 3.7 & 3.9 & 7.3 & 2.4 & 3.3 & 3.4 & 6.1 & 4.4 & 1.7 & 2.2 & 4.7 & 2.6 & 1.9 & 4.2 & 3.9 & 6.1 & 3.9 & 2.0 & 2.4 & 5.6 & 4.34 & 334.4 \\
granite-speech-3.3-2b & 3.6 & 5.6 & 4.9 & 9.8 & 2.1 & 4.0 & 5.5 & 6.0 & 5.5 & 3.0 & 3.0 & 6.1 & 5.5 & 2.2 & 9.5 & 4.0 & 22.4 & 5.7 & 2.2 & 3.5 & 5.0 & 10.48 & 948.2 \\
granite-speech-3.3-8b & 3.2 & 5.6 & 3.8 & 8.7 & 2.9 & 4.2 & 5.6 & 5.4 & 4.7 & 3.0 & 3.2 & 5.6 & 5.7 & 2.5 & 8.5 & 3.5 & 8.6 & 4.9 & 3.0 & 3.8 & 4.8 & 3.51 & 250.6 \\
\bottomrule
\end{tabular}
}
\end{landscape}

\section{Insertion rates and hallucination categories}
\label{app:insertion}

Table~\ref{tab:insertion_accent} reports insertion rates by accent on Common Voice~24.
Whisper-large-v3's Indian-accent insertion rate (9.62\%) is the most extreme value in the table, $6.3\times$ higher than Whisper-medium (1.53\%) and far above any Gen~3 model ($\leq$3.07\%).
This effect is accent-selective: Whisper-large-v3's insertion rates on US (0.96\%), Australian (0.90\%), and Canadian (0.52\%) accents are comparable to other models.

\begin{table}[t]
\centering
\caption{Insertion rate (\%) by accent across ASR models on Common Voice~24. Whisper-large-v3's Indian-accent insertion rate (9.62\%) is $6.3\times$ higher than Whisper-medium (1.53\%) and dramatically exceeds all Gen~3 models ($\leq$3.07\%).}
\label{tab:insertion_accent}
\resizebox{\textwidth}{!}{
\begin{tabular}{lrrrrrr}
\toprule
Model & US & England & Canada & Australia & Indian & African \\
\midrule
wav2vec2-large & 1.88 & 1.77 & 2.41 & 1.51 & 1.79 & 1.53 \\
whisper-small & 1.93 & 2.45 & 1.05 & 1.00 & 3.22 & 4.59 \\
whisper-medium & 1.30 & 1.16 & 0.84 & 0.80 & 1.53 & 1.15 \\
whisper-large-v3 & 0.96 & 0.79 & 0.52 & 0.90 & \textbf{9.62} & 3.63 \\
qwen3-asr-0.6b & 0.89 & 0.82 & 0.52 & 0.60 & 1.05 & 1.15 \\
qwen3-asr-1.7b & 0.57 & 0.45 & 1.78 & 0.60 & 0.68 & 0.96 \\
canary-qwen-2.5b & 0.72 & 0.58 & 1.88 & 0.40 & 0.72 & 0.77 \\
granite-speech-3.3-2b & 0.73 & 0.87 & 0.31 & 0.20 & 0.82 & 0.78 \\
granite-speech-3.3-8b & 1.74 & 0.90 & 0.63 & 0.50 & 3.07 & 1.53 \\
\bottomrule
\end{tabular}
}
\end{table}

Table~\ref{tab:indian_hallucination} decomposes insertions on Indian-accented    
  speech by hallucination type.                                                    
  Whisper-large-v3 is the only model with substantial repetition loops (43.0\%),   
  driven by autoregressive cycling on out-of-distribution acoustic input.          
  Granite-8B shows a secondary repetition mode (70.3\% of its 158 insertions),     
  likely amplified by Q-former compression.                                      
  All other models' insertions are predominantly content-classified (phonetic      
  mismatches scored as insertions by the alignment algorithm) rather than true   
 hallucinations.                                                                  
                                                                                 
  \begin{table}[h]
  \centering       
  \vspace{1em}
  \caption{Hallucination category breakdown on Indian-accented speech (Common      
  Voice~24, $n{=}511$ utterances, 5{,}154 reference words). Insertions classified
  as repetition (autoregressive loops), syntactic (function-word completions), or  
  content (semantic fabrications / phonetic mismatches).}                        
  \label{tab:indian_hallucination}
  \resizebox{\textwidth}{!}{
  \begin{tabular}{llrrrrrrr}                                                       
  \toprule                                                            
   & & & \multicolumn{3}{c}{Count} & \multicolumn{3}{c}{Percentage} \\             
  \cmidrule(lr){4-6} \cmidrule(lr){7-9}                                          
  Model & Gen & Ins.\ Total & Rep. & Syn. & Content & Rep.\% & Syn.\% & Content\%  
  \\                                                                             
  \midrule                                                                         
  Wav2Vec2-large        & 1 &  92 &   0 &  32 &  60 &  0.0 & 34.8 & 65.2 \\        
  Whisper-small         & 2 & 166 &   3 &  62 & 101 &  1.8 & 37.3 & 60.8 \\        
  Whisper-medium        & 2 &  79 &   0 &  30 &  49 &  0.0 & 38.0 & 62.0 \\        
  Whisper-large-v3      & 2 & \textbf{495} & \textbf{213} & \textbf{236} &  46 &   
  \textbf{43.0} & \textbf{47.7} &  9.3 \\                                          
  Qwen3-ASR-0.6B       & 3 &  54 &   0 &  17 &  37 &  0.0 & 31.5 & 68.5 \\         
  Qwen3-ASR-1.7B       & 3 &  35 &   0 &  11 &  24 &  0.0 & 31.4 & 68.6 \\         
  Canary-Qwen-2.5B     & 3 &  37 &   0 &  13 &  24 &  0.0 & 35.1 & 64.9 \\         
  Granite-Speech-2B     & 3 &  42 &   0 &  19 &  23 &  0.0 & 45.2 & 54.8 \\        
  Granite-Speech-8B     & 3 & 158 & 111 &  14 &  33 & 70.3 &  8.9 & 20.9 \\        
  \bottomrule                                                                      
  \end{tabular}                                                                    
  }                                                               
  \end{table}   

\begin{figure}[h]
\centering
\includegraphics[width=\textwidth]{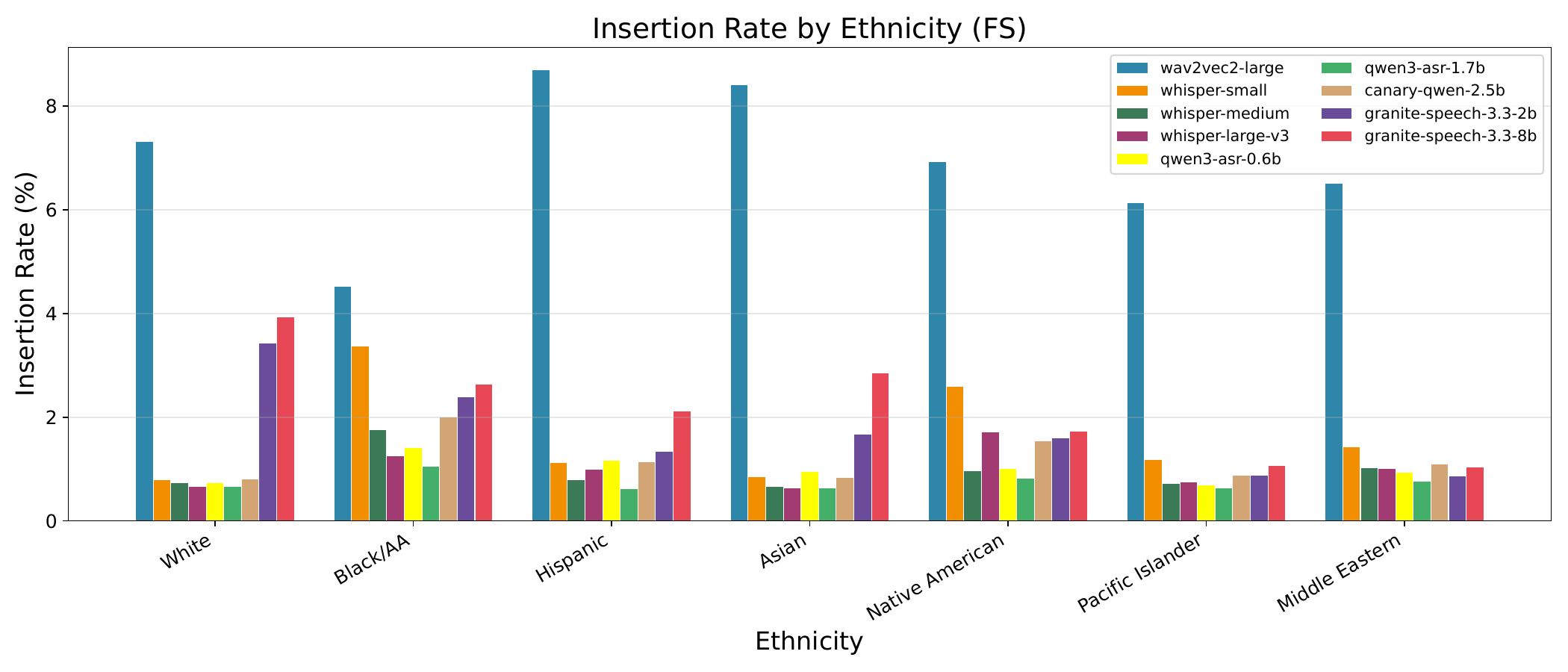}
\caption{Insertion rate (\%) by ethnicity on Fair-Speech.  Granite models show a reversed pattern: higher insertion rates for White than Black/AA speakers.}
\label{fig:insertion_ethnicity}
\end{figure}

Table~\ref{tab:insertion_ethnicity} and Figure~\ref{fig:insertion_ethnicity} report insertion rates by ethnicity on Fair-Speech.
Notably, Granite-2B and Granite-8B show \emph{higher} insertion rates for White speakers (3.42\%, 3.92\%) than for Black/AA speakers (2.40\%, 2.63\%), the opposite of all Whisper and other Gen~3 models (see \S\ref{sec:accent:type}). Wav2Vec2 also shows higher White than Black/AA insertion rates (7.30\% vs.\ 4.53\%), but this reflects its uniformly high error rates rather than a decoder-specific pattern.

\begin{table}[t]
\centering
\caption{Insertion Rate (\%) by Ethnicity}
\label{tab:insertion_ethnicity}
\resizebox{\textwidth}{!}{
\begin{tabular}{lrrrrrrr}
\toprule
Model & White & Black/AA & Hispanic & Asian & Native American & Pacific Islander & Middle Eastern \\
\midrule
wav2vec2-large & 7.30 & 4.53 & 8.70 & 8.41 & 6.93 & 6.14 & 6.51 \\
whisper-small & 0.78 & 3.37 & 1.13 & 0.85 & 2.60 & 1.19 & 1.43 \\
whisper-medium & 0.71 & 1.75 & 0.80 & 0.66 & 0.97 & 0.72 & 1.03 \\
whisper-large-v3 & 0.64 & 1.25 & 0.99 & 0.63 & 1.71 & 0.76 & 1.00 \\
qwen3-asr-0.6b & 0.71 & 1.41 & 1.16 & 0.95 & 1.02 & 0.69 & 0.93 \\
qwen3-asr-1.7b & 0.65 & 1.05 & 0.62 & 0.63 & 0.83 & 0.63 & 0.76 \\
canary-qwen-2.5b & 0.79 & 2.00 & 1.15 & 0.83 & 1.54 & 0.88 & 1.10 \\
granite-speech-3.3-2b & 3.42 & 2.40 & 1.34 & 1.68 & 1.62 & 0.88 & 0.87 \\
granite-speech-3.3-8b & 3.92 & 2.63 & 2.12 & 2.85 & 1.73 & 1.07 & 1.04 \\
\bottomrule
\end{tabular}
}
\end{table}

\begin{figure}[h]
\centering
\includegraphics[width=\textwidth]{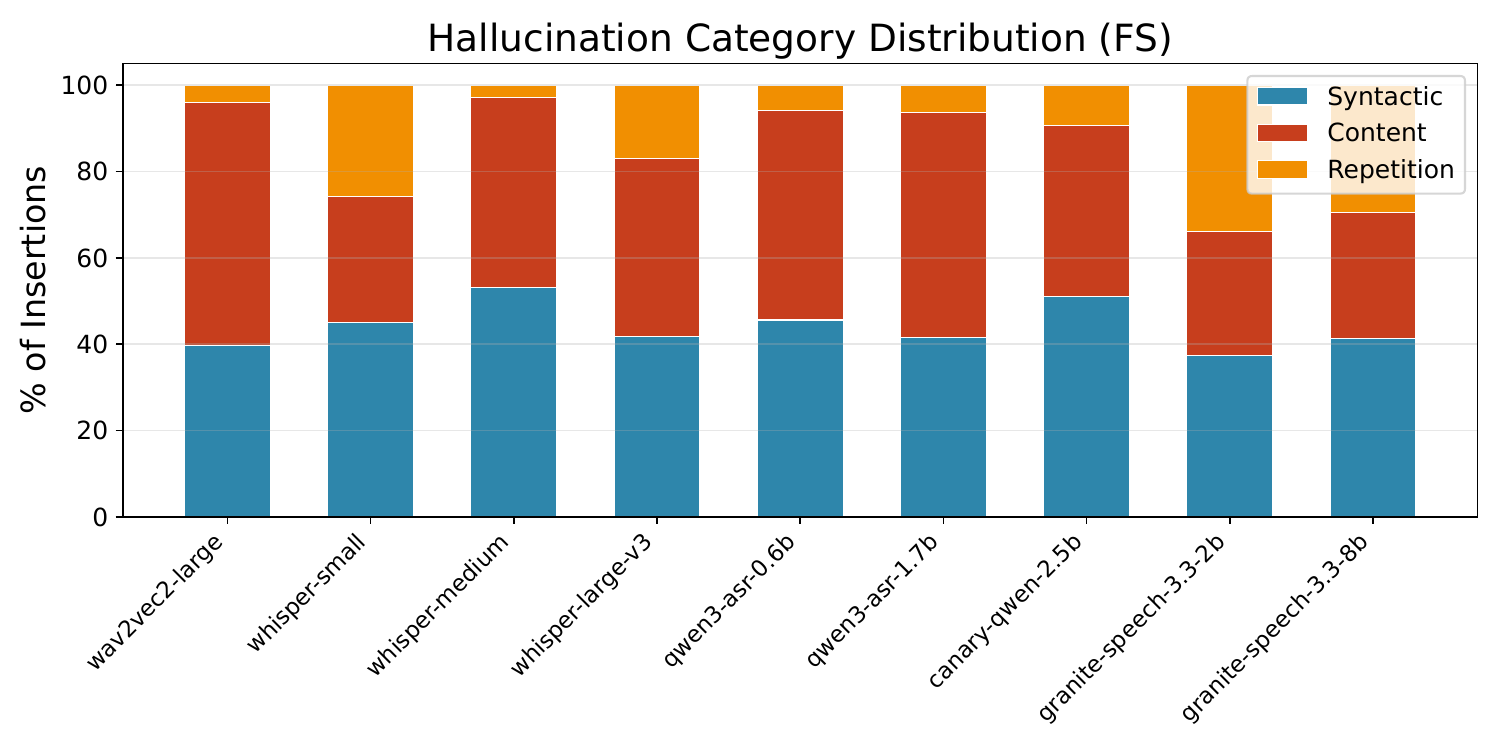}
\caption{Hallucination category distribution on Fair-Speech.
  Whisper-large-v3's insertions are dominated by syntactic completions and repetitions; Gen~3 models have proportionally more content-classified insertions but far fewer total insertions.}
\label{fig:hallucination_categories_fs}
\end{figure}

Figure~\ref{fig:hallucination_categories_fs} shows the hallucination category distribution on Fair-Speech, following the same methodology as the Common Voice distribution (Figure~\ref{fig:hallucination_categories_cv} in the main text).

\section{Inference configuration}
\label{app:inference}

Table~\ref{tab:inference_config} details the inference configuration for each model.
All models use greedy decoding (no beam search, no sampling) to ensure deterministic and reproducible results.
Text normalization is applied identically to both references and hypotheses using Whisper's \texttt{EnglishTextNormalizer}, which handles case folding, punctuation removal, number normalization (``six'' $\leftrightarrow$ ``6''), and contraction expansion (``she'll'' $\to$ ``she will'').

\begin{table}[h]
\centering
\caption{Inference configuration for all models.
  All use greedy decoding with identical text normalization.}
\label{tab:inference_config}
\resizebox{\textwidth}{!}{
\begin{tabular}{lllcl}
\toprule
Model & HuggingFace ID & Decoding & Max Tokens & Prompt / Notes \\
\midrule
Wav2Vec2-large
  & \texttt{facebook/wav2vec2-large-960h-lv60-self}
  & CTC argmax
  & n/a
  & No prompt; CTC output \\
Whisper-small
  & \texttt{openai/whisper-small}
  & Greedy
  & 440
  & \texttt{language="en", task="transcribe"} \\
Whisper-medium
  & \texttt{openai/whisper-medium}
  & Greedy
  & 440
  & \texttt{language="en", task="transcribe"} \\
Whisper-large-v3
  & \texttt{openai/whisper-large-v3}
  & Greedy
  & 440
  & \texttt{language="en", task="transcribe"} \\
Qwen3-ASR-0.6B
  & \texttt{Qwen/Qwen3-ASR-0.6B}
  & Greedy
  & 256
  & \texttt{language="English"} via \texttt{qwen-asr} \\
Qwen3-ASR-1.7B
  & \texttt{Qwen/Qwen3-ASR-1.7B}
  & Greedy
  & 256
  & \texttt{language="English"} via \texttt{qwen-asr} \\
Canary-Qwen-2.5B
  & \texttt{nvidia/canary-qwen-2.5b}
  & Greedy
  & 256
  & NeMo SALM API; audio locator tag prompt \\
Granite-Speech-2B
  & \texttt{ibm-granite/granite-speech-3.3-2b}
  & Greedy
  & 256
  & Chat template; \texttt{do\_sample=False, num\_beams=1} \\
Granite-Speech-8B
  & \texttt{ibm-granite/granite-speech-3.3-8b}
  & Greedy
  & 256
  & Chat template; \texttt{do\_sample=False, num\_beams=1} \\
\bottomrule
\end{tabular}
}
\end{table}

\paragraph{Granite prompt template.}
Granite-Speech models use a chat-style prompt to constrain output:
\begin{quote}
\small
\textbf{System:} ``You are a speech transcription system. Output ONLY the exact transcription of the audio. Do not add any commentary, explanation, or formatting.'' \\
\textbf{User:} ``\texttt{<|audio|>}Transcribe this audio exactly.''
\end{quote}
\noindent A post-processing step extracts the transcription from any residual conversational wrapping.

\paragraph{Audio preprocessing.}
All audio is resampled to 16\,kHz mono before inference.
Whisper models use log-mel spectrogram features computed by the Whisper processor (80-dimensional for small/medium, 128-dimensional for large-v3).
Wav2Vec2 operates on raw waveforms.
Gen~3 models use their respective audio encoders (Qwen3: direct projection; Granite: Conformer + Q-former; Canary: FastConformer).

\section{Bootstrap confidence intervals}
\label{app:bootstrap}

Tables~\ref{tab:ci_ethnicity} and~\ref{tab:ci_accent} report 95\% bootstrap confidence intervals (200 resamples, utterance-level) for group-level WER on the two primary demographic axes.
Non-overlapping CIs between groups indicate statistically significant differences at the 95\% level.

\begin{table}[htp]
\centering
\begin{minipage}{\textwidth}
\caption{Bootstrap 95\% CIs for WER by ethnicity on Fair-Speech.
  Black/AA CIs do not overlap with White CIs for 8 of 9 models.}
\label{tab:ci_ethnicity}
\resizebox{\linewidth}{!}{
\begin{tabular}{lccccc}
\toprule
Model & Black/AA [95\% CI] & White [95\% CI] & Asian [95\% CI] & Hispanic [95\% CI] & Overlap? \\
\midrule
Wav2Vec2-large     & 41.59 [40.87, 42.26] & 26.25 [25.47, 27.18] & 28.74 [27.86, 29.58] & 30.17 [29.13, 31.28] & No \\
Whisper-small      & 18.57 [17.38, 20.04] & 10.48 [9.14, 11.63]  & 4.59 [4.21, 5.13]    & 7.60 [6.25, 9.09]    & No \\
Whisper-medium     & 12.42 [11.83, 12.92] & 9.85 [8.58, 11.13]   & 3.78 [3.34, 4.53]    & 6.24 [4.96, 7.49]    & No \\
Whisper-large-v3   & 10.25 [9.61, 10.81]  & 9.42 [8.21, 10.64]   & 3.27 [2.94, 3.66]    & 5.56 [4.54, 6.52]    & \textbf{Yes} \\
Qwen3-ASR-0.6B    & 11.01 [10.69, 11.33] & 3.00 [2.78, 3.29]    & 3.46 [3.19, 3.78]    & 3.84 [3.56, 4.20]    & No \\
Qwen3-ASR-1.7B    & 8.45 [8.13, 8.70]    & 2.79 [2.57, 3.05]    & 2.81 [2.59, 3.05]    & 2.93 [2.70, 3.23]    & No \\
Canary-Qwen-2.5B  & 11.65 [11.20, 12.04] & 4.90 [4.39, 5.46]    & 3.36 [3.11, 3.62]    & 3.81 [3.46, 4.28]    & No \\
Granite-Speech-2B  & 13.41 [12.74, 14.30] & 9.96 [8.43, 11.49]   & 5.29 [4.94, 5.80]    & 5.62 [5.08, 6.10]    & No \\
Granite-Speech-8B  & 12.09 [11.35, 12.87] & 7.57 [6.18, 8.93]    & 6.06 [5.05, 7.42]    & 5.46 [4.96, 5.96]    & No \\
\bottomrule
\end{tabular}
}
\end{minipage}

\vspace{2em}

\begin{minipage}{\textwidth}
\caption{Bootstrap CIs for accent on CV~24.
  Whisper-large-v3's wide Indian-accent CI [9.46, 43.53] reflects variance from hallucination episodes.}
\label{tab:ci_accent}
\resizebox{\linewidth}{!}{
\begin{tabular}{lcccc}
\toprule
Model & US [95\% CI] & Indian [95\% CI] & African [95\% CI] & England [95\% CI] \\
\midrule
Wav2Vec2-large     & 16.38 [15.03, 17.51] & 25.17 [23.25, 27.40] & 23.33 [16.92, 30.79] & 16.45 [14.85, 18.31] \\
Whisper-small      & 11.48 [10.44, 12.78] & 17.58 [15.86, 19.54] & 20.27 [12.16, 29.07] & 13.08 [10.90, 15.11] \\
Whisper-medium     & 8.79 [7.95, 9.89]    & 13.19 [11.75, 14.59] & 11.47 [6.89, 16.21]  & 9.44 [7.94, 11.31]  \\
Whisper-large-v3   & 7.45 [6.69, 8.54]    & 18.99 [9.46, 43.53]  & 13.00 [6.51, 21.81]  & 7.67 [6.45, 9.23]   \\
Qwen3-ASR-0.6B    & 7.53 [6.75, 8.47]    & 11.08 [9.69, 12.50]  & 9.56 [5.65, 14.27]   & 8.07 [6.84, 9.52]   \\
Qwen3-ASR-1.7B    & 5.76 [5.15, 6.56]    & 7.78 [6.69, 9.07]    & 8.03 [4.83, 12.39]   & 6.72 [5.57, 8.02]   \\
Canary-Qwen-2.5B  & 5.84 [5.09, 6.57]    & 7.17 [6.15, 8.24]    & 8.03 [4.71, 12.48]   & 6.27 [5.20, 7.73]   \\
Granite-Speech-2B  & 8.35 [7.24, 9.41]    & 10.36 [8.98, 11.89]  & 9.98 [5.29, 16.22]   & 9.04 [7.32, 11.10]  \\
Granite-Speech-8B  & 10.63 [8.43, 13.56]  & 12.87 [9.15, 18.02]  & 16.06 [8.65, 23.93]  & 9.44 [7.48, 11.64]  \\
\bottomrule
\end{tabular}
}
\end{minipage}
\end{table}

\section{WER degradation curves}
\label{app:perturbation_wer}

\begin{figure}[ht]
\centering
\includegraphics[width=\textwidth]{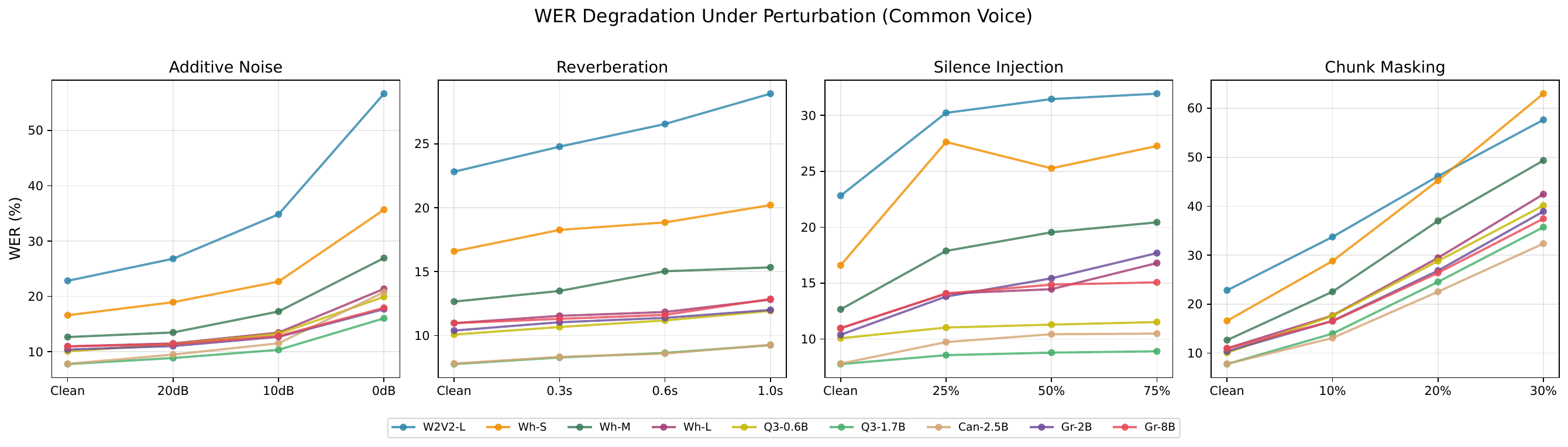}
\caption{WER degradation curves on Common Voice~24.  Patterns mirror Fair-Speech: Gen~3 models degrade most gracefully, masking is harshest, reverb is mildest.}
\label{fig:wer_curves_cv}
\end{figure}

Figure~\ref{fig:wer_curves_cv} shows overall WER degradation across 12 perturbation conditions for all nine models on Common Voice~24, following the Fair-Speech results shown in the main text (Figure~\ref{fig:wer_curves_fs}).
Masking is consistently the harshest perturbation; reverberation is the mildest for all models with language priors.
Gen~3 models (especially Qwen3) degrade most gracefully.

\section{Additional perturbation results}
\label{app:perturbation_full}

\begin{figure}[ht]
\centering
\begin{subfigure}[b]{0.49\textwidth}
\centering
\includegraphics[width=\textwidth]{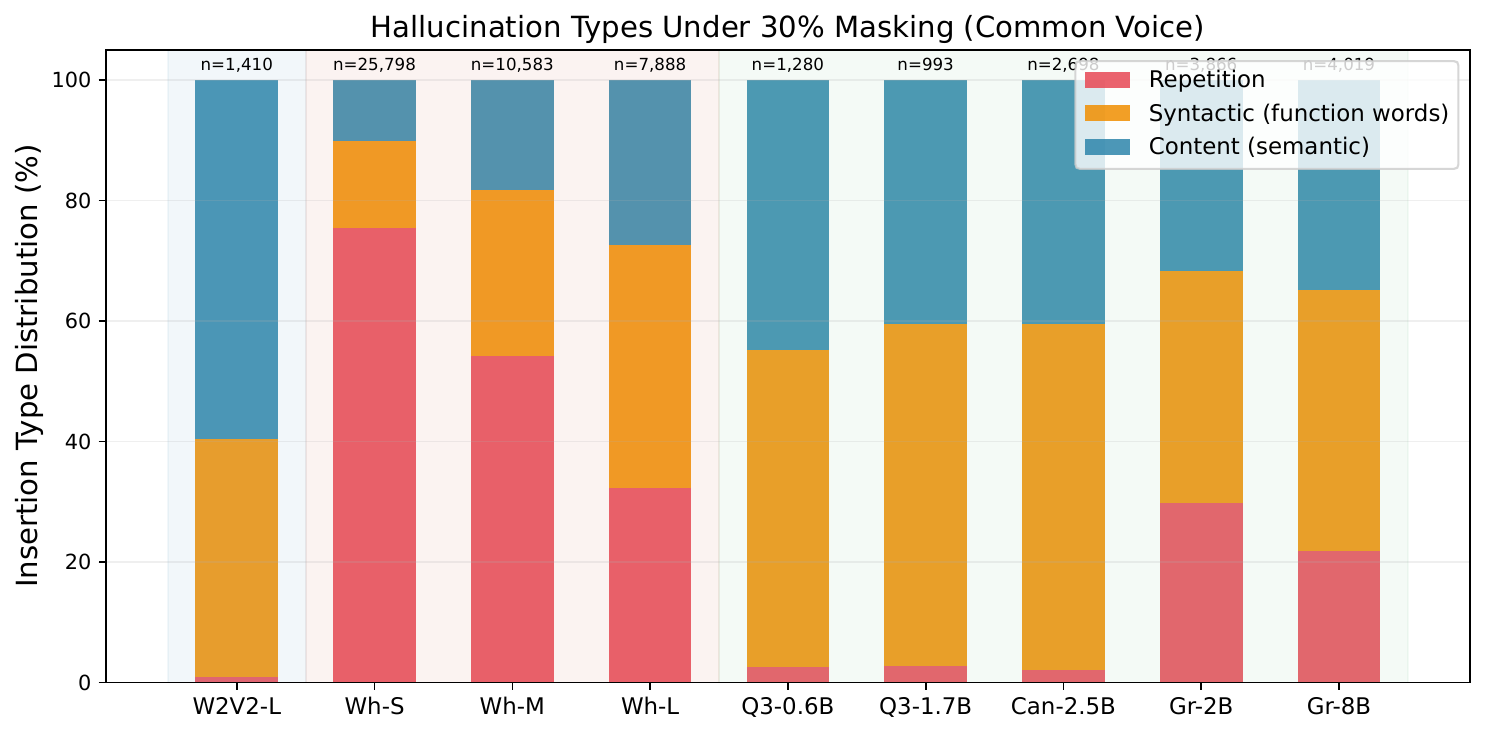}
\caption{Masking on Common Voice~24.}
\label{fig:hallucination_masking_cv}
\end{subfigure}
\hfill
\begin{subfigure}[b]{0.49\textwidth}
\centering
\includegraphics[width=\textwidth]{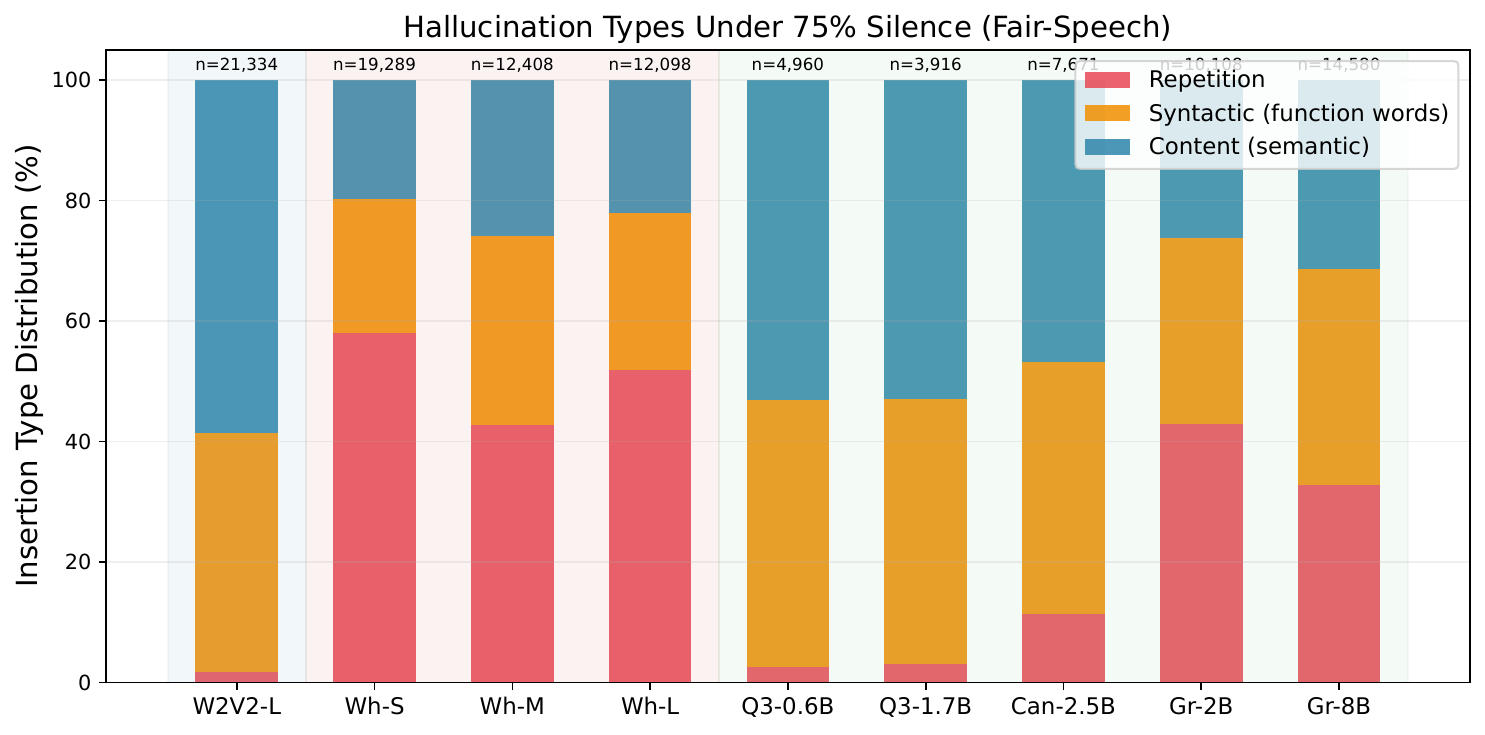}
\caption{Silence on Fair-Speech.}
\label{fig:hallucination_silence_fs}
\end{subfigure}

\vspace{0.5cm}
\begin{subfigure}[b]{0.49\textwidth}
\centering
\includegraphics[width=\textwidth]{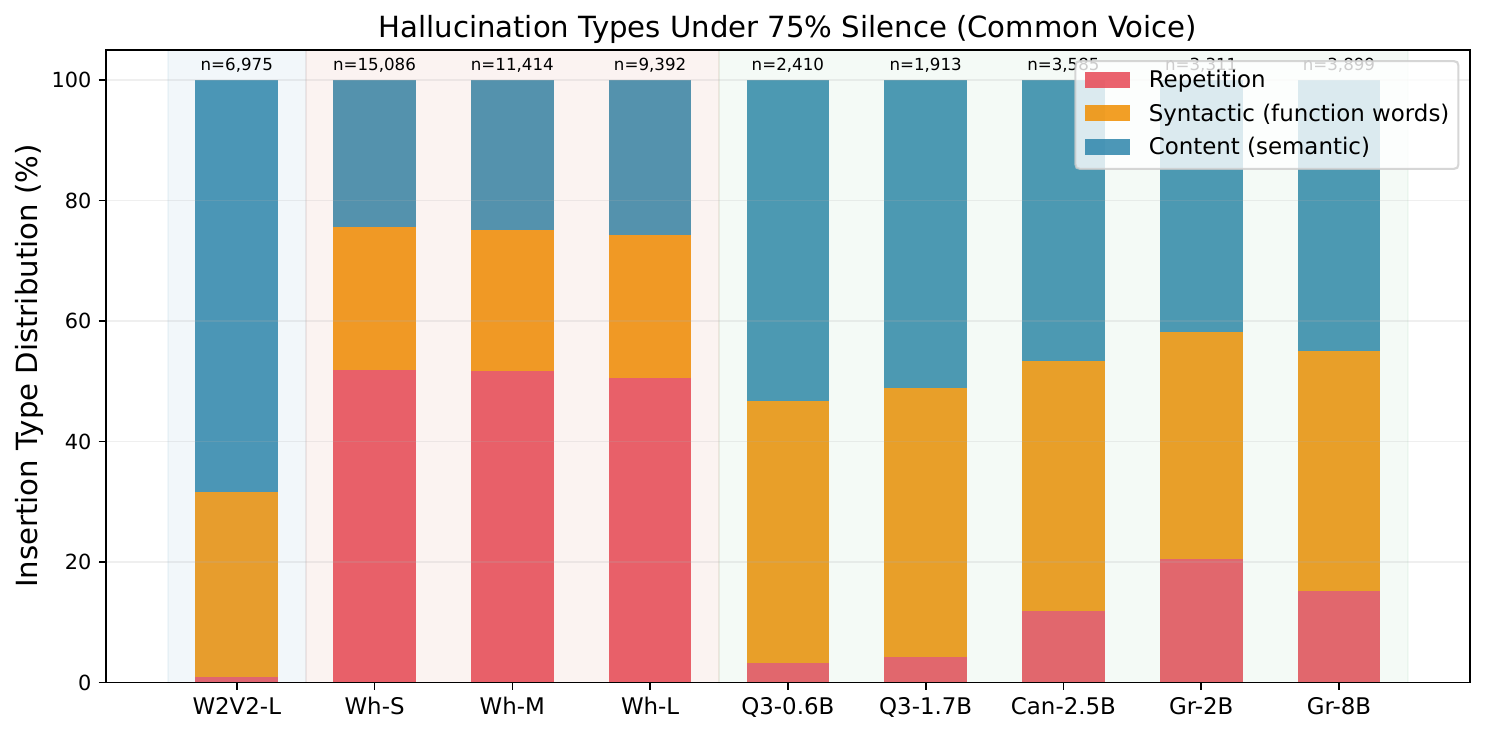}
\caption{Silence on Common Voice~24.}
\label{fig:hallucination_silence_cv}
\end{subfigure}
\caption{Hallucination type distributions under masking and silence injection.}
\end{figure}

\paragraph{Hallucination types under perturbation.}
Figures~\ref{fig:hallucination_masking_cv}-\ref{fig:hallucination_silence_cv} show additional hallucination classifications.
The main text (Figure~\ref{fig:hallucination_masking_fs}) details the distribution under masking on Fair-Speech.
Under masking on Common Voice, we observe similar dynamics.

\vspace{1em}
The hallucination distributions in Figures~\ref{fig:hallucination_silence_fs} and~\ref{fig:hallucination_silence_cv} confirm that signal dropout, whether via masking or silence, triggers fundamentally different failure modes across generations. While Whisper-small remains dominated by repetition loops ($>50\%$), Gen~3 explicit-LLM decoders shift their hallucination profile depending on the perturbation: under masking (Figure~\ref{fig:hallucination_masking_cv}) they predominantly produce benign syntactic completions, but under severe silence injection they produce primarily content substitutions. Crucially, explicitly-pretrained LLM priors prevent the catastrophic repetition loops seen in implicit-LM systems, maintaining higher syntactic integrity even when acoustic information is entirely removed.

\paragraph{Pareto frontier under perturbation.}

\begin{figure}[ht]
\centering
\includegraphics[width=0.6\textwidth]{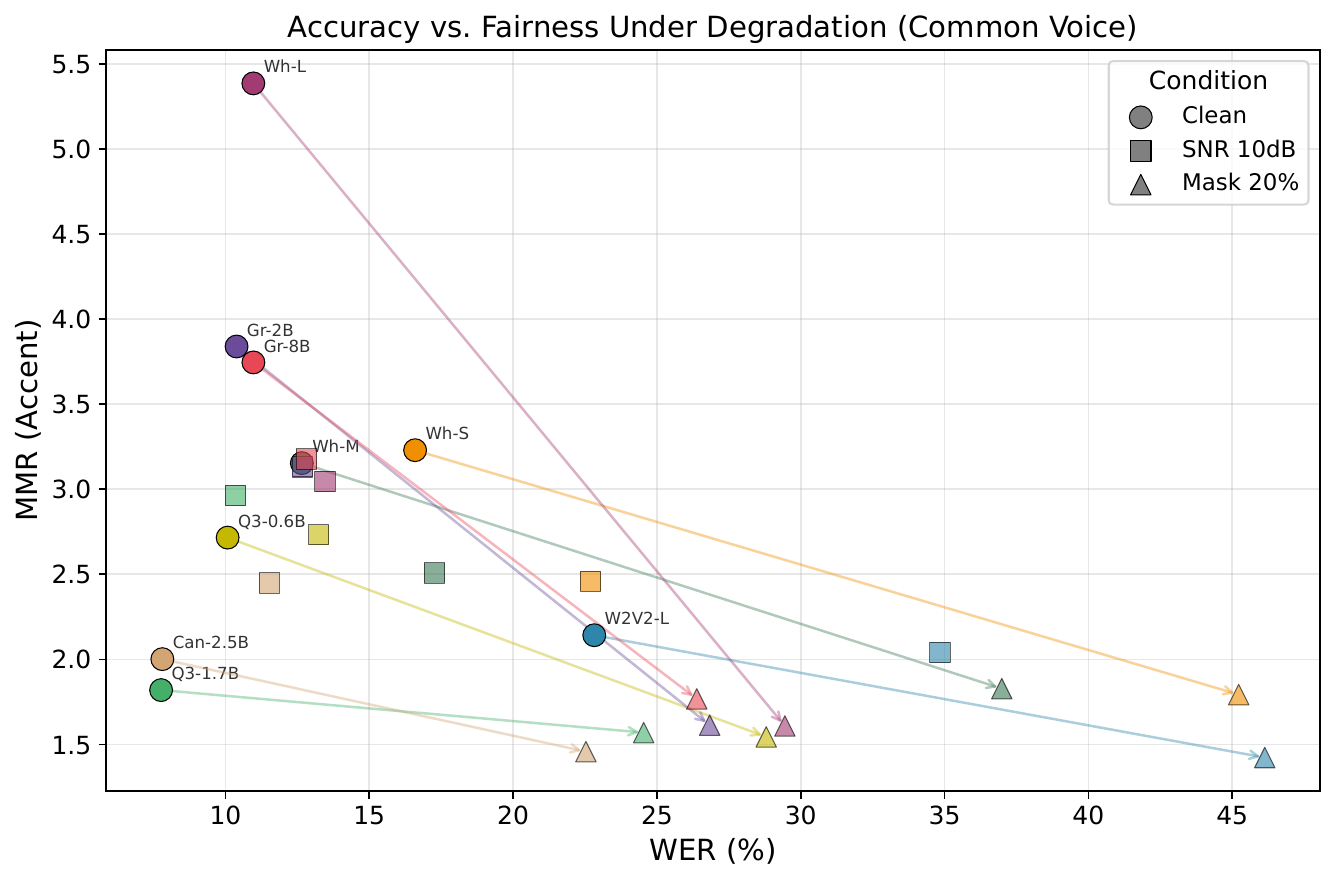}
\caption{Accuracy vs.\ accent fairness under perturbation on Common Voice~24.}
\label{fig:perturbation_pareto_accent}
\end{figure}

We provide the accuracy-fairness Pareto frontier for Common Voice accents in Figure~\ref{fig:perturbation_pareto_accent}. This mirrors the ethnicity findings from the main text (Figure~\ref{fig:pareto}): as models are stressed by masking (triangles), they move toward a ``degenerate frontier'' of low accuracy and low disparity. This visualization is essential for demonstrating that the ``degradation as equalizer'' effect is a universal property of autoregressive decoders rather than a dataset-specific artifact of Fair-Speech.

\begin{figure}[ht]
\centering
\begin{subfigure}[t]{0.49\textwidth}
\centering
\includegraphics[width=\textwidth]{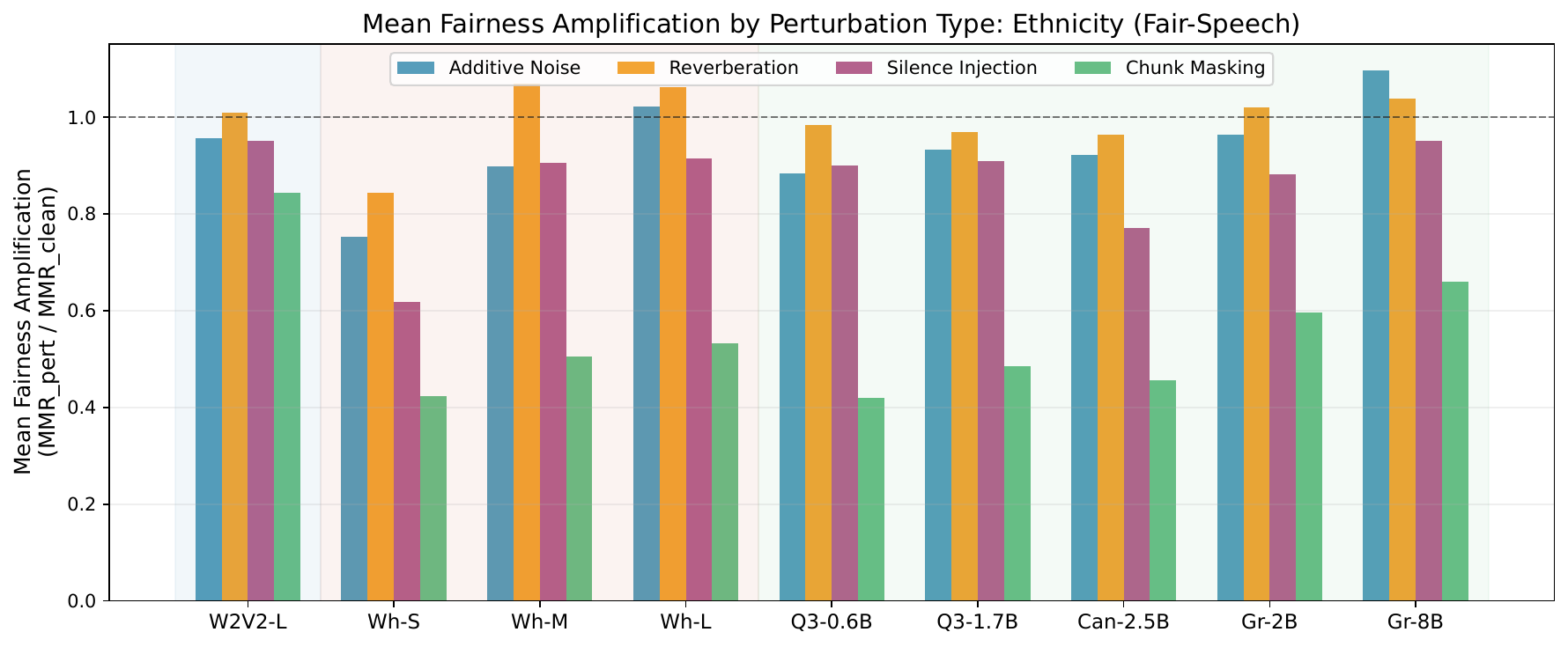}
\caption{Ethnicity on Fair-Speech.}
\label{fig:amplification_by_type_ethnicity}
\end{subfigure}
\hfill
\begin{subfigure}[t]{0.49\textwidth}
\centering
\includegraphics[width=\textwidth]{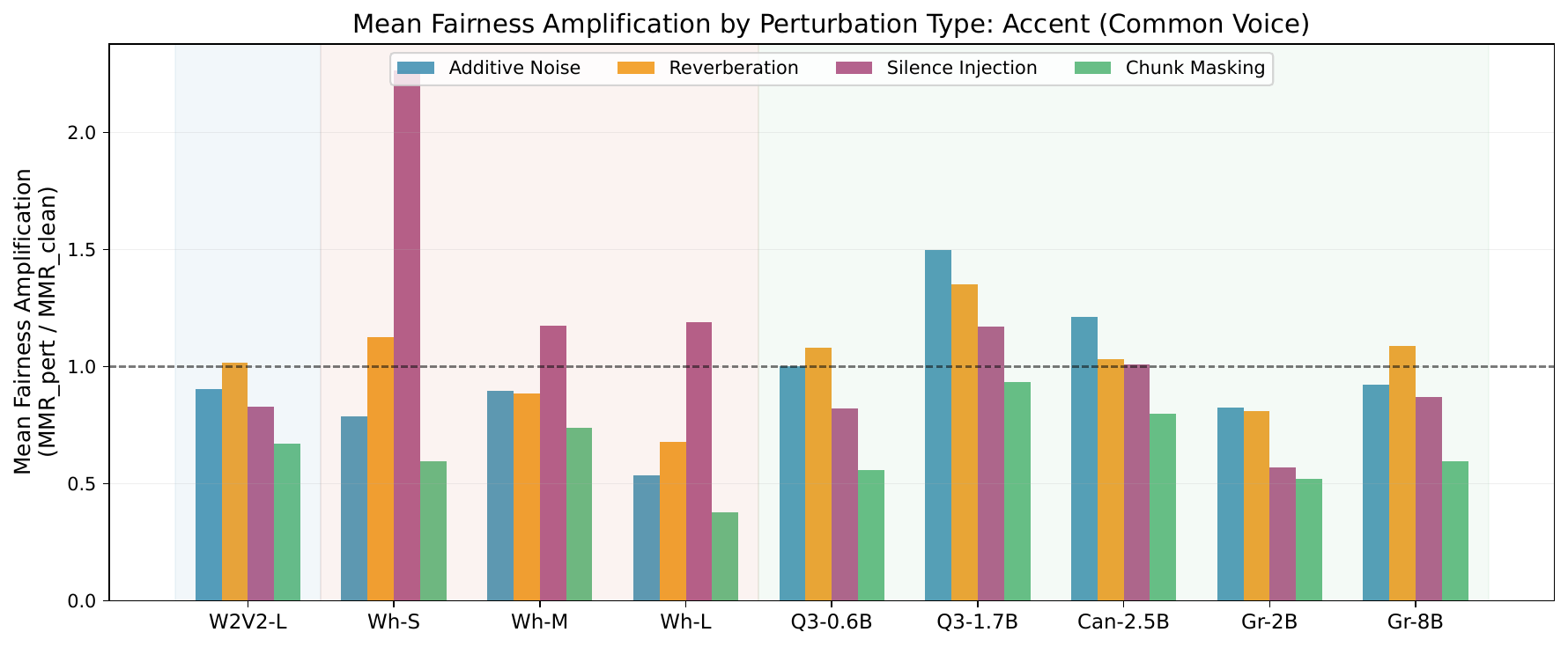}
\caption{Accent on Common Voice~24.}
\label{fig:amplification_by_type_accent}
\end{subfigure}
\caption{Mean fairness gap amplification grouped by perturbation type.}
\end{figure}

\paragraph{Amplification by perturbation type.}
Finally, Figures~\ref{fig:amplification_by_type_ethnicity} and~\ref{fig:amplification_by_type_accent} provide a high-level summary of the fairness gap amplification ratio ($\alpha$) across all 216 inference runs, grouped by perturbation type. These plots highlight a key architectural takeaway: while Chunk Masking universally compresses gaps ($\alpha < 1$), Silence Injection is uniquely hazardous for Whisper-small's accent fairness, and Additive Noise disproportionately stresses the Qwen3-1.7B accent frontier. This suggests that ``robustness'' is not a monolithic trait but is highly sensitive to the specific type of signal corruption.

\paragraph{Accent WER under perturbation.}
Table~\ref{tab:accent_perturbation} provides per-accent WER and insertion rates for the two key amplification cases discussed in \S\ref{sec:perturbation:amplification}.
Under SNR\,{=}\,10\,dB, Qwen3-1.7B's Indian-accent WER rises disproportionately while Canadian-accent WER barely changes, driving $\alpha$\,{=}\,1.63.
Under 75\% silence, Whisper-large-v3's worst-affected group shifts from Indian to England: Indian-accent WER and insertion rate both \emph{decrease}, while England-accent insertion rate rises from 0.8\% to 22.1\%.

\begin{table}[htp]
\centering
\vspace{1em}
\caption{Per-accent WER (\%) and insertion rate (\%) on Common Voice~24 under key perturbation conditions. Top: Qwen3-1.7B under SNR\,{=}\,10\,dB noise. Bottom: Whisper-large-v3 under 75\% silence injection. These conditions produce the strongest fairness gap amplification ($\alpha$\,{=}\,1.63 and 1.38, respectively; \S\ref{sec:perturbation:amplification}).}
\label{tab:accent_perturbation}
\resizebox{\textwidth}{!}{
\begin{tabular}{llcccccccc}
\toprule
& & African & Australia & Canada & England & Indian & US & MMR & $\alpha$ \\
& & ($n{=}51$) & ($n{=}98$) & ($n{=}99$) & ($n{=}381$) & ($n{=}511$) & ($n{=}1193$) & & \\
\midrule
\multicolumn{9}{l}{\textbf{Qwen3-1.7B} (clean accent MMR\,{=}\,1.82)} \\
\quad WER & Clean & 8.0 & 4.4 & 4.6 & 6.7 & 7.8 & 5.8 & 1.82 & n/a \\
\quad WER & SNR 10\,dB & 11.5 & 4.4 & 3.9 & 8.0 & 11.3 & 7.4 & 2.96 & 1.63 \\
\midrule
\multicolumn{9}{l}{\textbf{Whisper-large-v3} (clean accent MMR\,{=}\,5.38)} \\
\quad WER & Clean & 13.0 & 5.2 & 3.6 & 7.7 & 19.2 & 7.5 & 5.38 & n/a \\
\quad WER & Silence 75\% & 13.6 & 7.1 & 4.1 & 30.4 & 14.5 & 11.3 & 7.45 & 1.38 \\
\quad Ins.\ rate & Clean & 3.6 & 0.9 & 0.5 & 0.8 & 9.6 & 1.0 & & \\
\quad Ins.\ rate & Silence 75\% & 2.9 & 1.2 & 0.6 & 22.1 & 2.6 & 3.6 & & \\
\bottomrule
\end{tabular}
}
\end{table}

\section{LLM Usage Disclosure}
\label{app:llm_usage}

The core ideas, experimental design, data selection and analysis in this work were solely conducted by the authors. AI assistance was used to facilitate code implementation, fix formatting, and refine the prose for clarity and flow.

\end{document}